\pdfoutput=1
\documentclass{llncs}

\usepackage{amsmath,amsfonts,amssymb,mathrsfs}
\usepackage{psfrag,graphicx,epsfig,epsf}
\usepackage[ruled,linesnumbered, noend]{algorithm2e}
\usepackage{fancyhdr}
\usepackage{wrapfig}
\usepackage{subfigure}
\usepackage{mathtools}
\usepackage{url}
\usepackage{color}
\usepackage{verbatim}
\usepackage{xspace}

\newcommand\permu[2][^n]{\prescript{#1\mkern-2.5mu}{}P_{#2}}

\begin{document}

\title{Fast, High-Quality Dual-Arm Rearrangement in Synchronous, Monotone Tabletop Setups}
%\titlerunning{Fast Methods for Synchronized Tabletop Dual-Arm Rearragement}
\author{Rahul Shome$^1$ \and Kiril Solovey$^2$ \and Jingjin Yu$^1$ \and Kostas Bekris$^1$ \and Dan Halperin$^2$}
%\authorrunning{Shome et al.}
\institute{$^1$Rutgers University, NJ, USA and $^2$Tel Aviv University, Israel}

\maketitle
\newcommand{\danh}[2][1=]{\todo[linecolor=blue,
			backgroundcolor=blue!5,bordercolor=black,#1]{DH:#2}}
\newcommand{\kb}[2][1=]{\todo[linecolor=green,
			backgroundcolor=green!5,bordercolor=black,#1]{KB:#2}}
\newcommand{\ks}[2][1=]{\todo[linecolor=red,
			backgroundcolor=red!5,bordercolor=black,#1]{KS:#2}}
\newcommand{\rs}[2][1=]{\todo[linecolor=orange,
			backgroundcolor=orange!10,bordercolor=black,#1]{RS:#2}}
\newcommand{\jy}[2][1=]{\todo[linecolor=black,
			backgroundcolor=black!5,bordercolor=black,#1]{JJ:#2}}

%%% Mathematical Definitions
\newcommand{\reals}{\mathbb{R}}
\newcommand{\integers}{\mathbb{Z}}

%%% Definitions for Workspace, Objects, Manipulator
\newcommand{\Wspace}{\mathcal{W}}
\newcommand{\Objects}{\mathcal{O}}
\newcommand{\Manip}{\mathcal{M}}
\newcommand{\nobj}{k}

%% Definitions for Object stuff
\newcommand{\Pspace}{\mathcal{P}}
\newcommand{\Pstable}{\mathcal{P}^s}
\newcommand{\pose}{p}
\newcommand{\GeomObj}{\mathcal{WO}}
\newcommand{\Arrange}{\mathcal{A}}
\newcommand{\Pumped}{\mathcal{A^P}}
\newcommand{\pumpedarr}{\alpha^{\mathcal{P}}}

%% Definitions for Manipulator stuff
\newcommand{\Qspace}{\mathcal{Q}}
\newcommand{\GeomManip}{\mathcal{WM}}

%% Definitions for problem and state space
\newcommand{\Tspace}{\mathbb{T}} 
\newcommand{\Xspace}{\mathbb{X}}
\newcommand{\paths}{\Pi}

%% Manipulation roadmap definition
\newcommand{\roadmap}{\mathcal{R}}
\newcommand{\graph}{\mathcal{G}}
\newcommand{\nodes}{\mathcal{V}}
\newcommand{\node}{{v}}
\newcommand{\edges}{\mathcal{E}}
\newcommand{\edge}{{e}}
\newcommand{\prmstar}{{\tt PRM$^*$}}

\newcommand{\rpg}{${\tt RPG}$}

\newcommand{\local}{\mathcal{L}}

\newcommand{\prm}{{\tt PRM}}
\newcommand{\kprmstar}{{\tt k-PRM$^*$}}
\newcommand{\rrt}{{\tt RRT}}
\newcommand{\rrtdrain}{{\tt RRT-Drain}}
\newcommand{\rrg}{{\tt RRG}}
\newcommand{\est}{{\tt EST}}
\newcommand{\rrtstar}{{\tt RRT$^*$}}
\newcommand{\srrt}{{\tt RDG}}
\newcommand{\bvp}{{\tt BVP}}
\newcommand{\rdg}{{\tt RDG}}
\newcommand{\lrg}{{\tt LRG}}
\newcommand{\alg}{{\tt ALG}}
\newcommand{\upump}{{\tt UPUMP}}
\newcommand{\prxpump}{{\tt RPG}}
\newcommand{\fixed}{{\tt Fixed}-$\alpha$-\rdg}
\newcommand{\nrob}{k}
\newcommand{\cons}{K}

\newcommand{\frnodes}{V_f}
\newcommand{\frnode}{v_f}
\newcommand{\grnodes}{V_g}
\newcommand{\grnode}{v_g}
\newcommand{\fredges}{E_f}
\newcommand{\fredge}{e_f}
\newcommand{\gredges}{E_g}
\newcommand{\gredge}{e_g}
\newcommand{\kedges}{E_{\cons}}
\newcommand{\kedge}{e_{\cons}}
\newcommand{\safe}{q_s^{\mathcal{M}}}
\newcommand{\hedges}{E_H}
\newcommand{\hedge}{e_H}
\newcommand{\hnodes}{V_H}
\newcommand{\hnode}{v_H}
\newcommand{\hgraph}{H}
\newcommand{\nblank}{b}
\newcommand{\config}{C}
\newcommand{\cquery}{\mathbb{C}}
\newcommand{\pumped}{P}
\newcommand{\pumpedgraph}{\mathcal{G}_P}
\newcommand{\pnodes}{V_P}
\newcommand{\pnode}{v_P}
\newcommand{\pedges}{E_P}
\newcommand{\pedge}{e_P}
\newcommand{\signs}{\Sigma}
\newcommand{\sign}{\sigma}
\newcommand{\gsign}{\sigma_{\pumpedgraph}}
\newcommand{\cedges}{E_c}
\newcommand{\constraints}{\tt c}

\newenvironment{myitem}{\begin{list}{$\bullet$}
{\setlength{\itemsep}{-0pt}
\setlength{\topsep}{0pt}
\setlength{\labelwidth}{0pt}
\setlength{\leftmargin}{10pt}
\setlength{\parsep}{-0pt}
\setlength{\itemsep}{0pt}
\setlength{\partopsep}{0pt}}}%
{\end{list}}

%\newtheorem{theorem}{Theorem}
%\newtheorem{definition}[theorem]{Definition}
%\newtheorem{proposition}[theorem]{Proposition}
%\newtheorem{corollary}[theorem]{Corollary}
%\newtheorem{axiom}[theorem]{Axiom}
%\newtheorem{lemma}[theorem]{Lemma}
%\newtheorem{problem}{Problem}
%\newtheorem{prob}[theorem]{Problem}
%\newtheorem{conjecture}[theorem]{Conjecture}
%\newtheorem{obj}[theorem]{Objective}
%\newtheorem{prop}[theorem]{Property}
%\newtheorem{schedule}[theorem]{Schedule}

%\newtheorem{definition}{\bf Definition}
%\newtheorem{assumption}{\bf Assumption}
%\newtheorem{thm}{\bf Theorem}
%\newtheorem{requirement}{\bf Requirement}
%\newtheorem{lemmma}{\bf Lemma}
%\newtheorem{coro}{\bf Corollary}

%%%%%%%%%%%%%%%%%%%%%%%%%%%%%%%%%%%%
%% Nick Saving space
%%%%%%%%%%%%%%%%%%%%%%%%%%%%%%%%%%%%
% Space between figure and caption
%\setlength{\abovecaptionskip}{-2.5pt}
%\setlength{\belowcaptionskip}{-6pt}
%% Space between text and figs
%\setlength{\dbltextfloatsep}{2pt plus 1.0pt minus 1.0pt}
%\setlength{\textfloatsep}{2pt plus 1.0pt minus 1.0pt}
%\setlength{\intextsep}{2pt plus 1.0pt minus 1.0pt}
%% Space between equations and text
%\setlength{\belowdisplayskip}{0pt} \setlength{\belowdisplayshortskip}{2pt}
%\setlength{\abovedisplayskip}{0pt} \setlength{\abovedisplayshortskip}{2pt}

\newcommand{\dof}{{\tt DoF}}

\newcommand{\mam}{$\mathcal{G}_{\tt MAM}$}
\newcommand{\pr}{\ensuremath{\mathbb{P}}}

\newcommand{\rad}{\ensuremath{r(n)}}
\newcommand{\radstar}{\ensuremath{r^*(n)}}
\newcommand{\radi}{\ensuremath{r_i(n)}}
\newcommand{\radj}{\ensuremath{r_j(n)}}
\newcommand{\crossrad}{\ensuremath{r_R(n)}}
\newcommand{\crossradstar}{\ensuremath{r^*_R(n)}}
\newcommand{\impcrossrad}{\ensuremath{\hat r_R(n)}}
\newcommand{\allimpcrossrad}{\ensuremath{\hat r_{R}(n^R)}}
\newcommand{\ki}{\ensuremath{k_i(n)}}
\newcommand{\kj}{\ensuremath{k_j(n)}}

%% Manipulation roadmap definition
\newcommand{\mmgraph}{\ensuremath{\mathbb{G}}}
\newcommand{\mmgimp}{\hat\mmgraph}
\newcommand{\mmgexp}{\mmgraph}
\newcommand{\aograph}{\ensuremath{\mathbb{G}^{AO}}}
\newcommand{\tree}{\ensuremath{\mathbb{T} \ }}
\newcommand{\mmnodes}{\mathbb{\hat V}}
\newcommand{\mmedges}{\mathbb{\hat E}}
\newcommand{\mmnodestpprm}{\mathbb{V}_{\chi_i}}
\newcommand{\mmedgestpprm}{\mathbb{E}_{\chi_i}}
\newcommand{\mmnode}{\mathbb{\hat v}}
\newcommand{\mmedge}{\mathbb{\hat e}}
\newcommand{\sprmstar}{Soft-\ensuremath{ {\tt PRM} }}
\newcommand{\irs}{\ensuremath{ {\tt IRS} }}
\newcommand{\spars}{{\tt SPARS}}
\newcommand{\drrt}{\ensuremath{{\tt dRRT}}}
\newcommand{\drrtstar}{\ensuremath{{\tt dRRT^*}}}
\newcommand{\dadrrtstar}{\ensuremath{\tt da\_dRRT^*}}

\newcommand{\sig}{{\tt SIG}}
\newcommand{\rmaps}{\ensuremath{\mathfrak{R}}}

\newcommand{\mmprm}{\ensuremath{\text{Random-}{\tt MMP}}}
\newcommand{\astar}{{\ensuremath{\tt A^{\text *}}}}
\newcommand{\mstar}{{\tt M^{\text *}}}
\newcommand{\opens}{P_{Heap}}

\newcommand{\cost}{\textup{cost}}

\newcommand{\kiril}[1]{{\color{blue} \textbf{Kiril:} #1}}
\newcommand{\chups}[1]{{\color{green} \textbf{Chuples:} #1}}
\newcommand{\rahul}[1]{{\color{red} \textbf{Rahul:} #1}}

\newcommand{\T}{\mathcal{T}}

% Dual Arm
\newcommand{\leftrm}{\ensuremath{\mathbb{R}_{l}}  }
\newcommand{\rightrm}{\ensuremath{\mathbb{R}_{r}}  }
\newcommand{\leftmetric}{\ensuremath{\mathbb{P}_{l}}  }
\newcommand{\rightmetric}{\ensuremath{\mathbb{P}_{r}}  }
\newcommand{\cfull}{\ensuremath{\mathbb{C}_{{\rm full}}}  }
\newcommand{\cfree}{\ensuremath{\mathbb{C}_{{\rm free}}}  }
\newcommand{\cobs}{\ensuremath{\mathbb{C}_{{\rm obs}}}  }
\newcommand{\cleft}{\ensuremath{\mathbb{C}_{{l}}}  }
\newcommand{\cright}{\ensuremath{\mathbb{C}_{{r}}}  }
\newcommand{\cshared}{\ensuremath{\mathbb{C}_{{s}}}  }
\newcommand{\cgoal}{\ensuremath{q_{{\rm goal}}}  }
\newcommand{\cstart}{\ensuremath{q_{{\rm start}}}  }

\newcommand{\gimpleft}{\ensuremath{\hat\mmgraph_l}}
\newcommand{\gimpright}{\ensuremath{\hat\mmgraph_r}}

\newcommand{\xrand}{\ensuremath{x^{\textup{rand} \ }}}
\newcommand{\xnear}{\ensuremath{x^{\textup{near} \ }}}
\newcommand{\xnew}{\ensuremath{x^{\textup{n}} \ }}
\newcommand{\xlast}{\ensuremath{x^{\textup{last} \ }}}
\newcommand{\xparent}{\ensuremath{x^{\textup{best} \ }}}

\newcommand{\lr}{\ensuremath{\mathbb{R}_{ls}}}
\newcommand{\rr}{\ensuremath{\mathbb{R}_{sr}}}
\newcommand{\lp}{\ensuremath{\mathbb{P}_{l}}}
\newcommand{\rp}{\ensuremath{\mathbb{P}_{r}}}

\newcommand{\motoman}{{\tt Motoman}}
\newcommand{\baxter}{{\tt Baxter}}
\newcommand{\ao}{{\tt AO}}

\newcommand\inlineeqno{\stepcounter{equation}\ (\theequation)}

\newcommand{\chomp}{\ensuremath{\tt CHOMP } }

\newtheorem{assumption}{Assumption}

\newcommand{\W}{\mathcal W}
\newcommand\perm[2][\^n]{\prescript{#1\mkern-2.5mu}{}P\_{#2}}
\newcommand\comb[2][\^n]{\prescript{#1\mkern-0.5mu}{}C\_{#2}}
\newcommand{\objectset}{\mathcal{O}}
\newcommand{\object}{o}
\newcommand{\workspace}{\mathcal{W}}
\newcommand{\taskspace}{\mathcal{T}}
\newcommand{\arrangement}{A}
\newcommand{\oar}{p}
\newcommand{\manipulators}{\mathcal{M}}
\newcommand{\manipulator}{\mathit{m}}
\newcommand{\arm}{m}
\newcommand{\taskset}{\mathcal{T}}
\newcommand{\task}{\mathit{T}}
\newcommand{\sol}{\Pi}
\newcommand{\state}{q}

\newcommand{\Aspace}{\mathcal{A}}
\newcommand{\Afree}{\mathcal{A}_{\rm val}}
\newcommand{\ainit}{A_{\rm init}}
\newcommand{\atarget}{A_{\rm goal}}
\newcommand{\soma}{{\tt soma}}
\newcommand{\coma}{\ensuremath{{\omega}}}
\newcommand{\scoma}{\ensuremath{{{\Omega}}}}
\newcommand{\qset}{\mathcal{Q}}
\newcommand{\startq}{S}
\newcommand{\targetq}{T}

\newcommand{\act}{a}
\newcommand{\actset}{\mathbb{A}}
\newcommand{\trajset}{{\D}}
\newcommand{\moveset}{\bar{\mathcal{O}}}
\newcommand{\home}{Q}
\newcommand{\scomaset}{\{\scoma\}}
\newcommand{\tour}{{\Gamma}}
\newcommand{\tspgraph}{\graph_{\tour}}
\newcommand{\tspnodes}{\nodes_{\tour}}
\newcommand{\tspedges}{\edges_{\tour}}
\newcommand{\algo}{{\tt{TOM}}\xspace}
\newcommand{\kuka}{{\tt{Kuka }}}
\newcommand{\D}{D}
\newcommand{\sininv}{\sin^{-1}}
\newcommand{\cosinv}{\cos^{-1}}
\newcommand{\milp}{{\tt{MILP}}\xspace}
%%%%%%%%%%%%%%%%%%%%%%%%%%%%%%
%Caption model
\newcounter{model}
%\addtocounter{model}{1}
\newenvironment{model}
{\refstepcounter{model}}
{\begin{center}
\textbf{Model. }~\themodel
\end{center}
\medskip}
%%%%%%%%%%%%%%%%%%%%%%%%%%%%%%
\definecolor{darkgreen}{RGB}{30,150,30}
\newcommand{\commentdel}[1]{{\color{magenta}}}
 \newcommand{\commentadd}[1]{{#1}}

\newcommand\blfootnote[1]{%
  \begingroup
  \renewcommand\thefootnote{}\footnote{#1}%
  \addtocounter{footnote}{-1}%
  \endgroup
}

\begin{abstract}
Rearranging objects on a planar surface arises in a variety of
robotic applications, such as product packaging.  Using two arms can improve
efficiency but introduces new computational challenges. This paper
studies the structure of dual-arm rearrangement for synchronous,
monotone tabletop setups and develops an optimal mixed integer model. It then
describes an efficient and scalable algorithm, which first minimizes
the cost of object transfers and then of moves between objects.
This is motivated by the fact that, asymptotically, object transfers
dominate the cost of solutions. Moreover, a lazy strategy minimizes
the number of motion planning calls and results in significant
speedups.  Theoretical arguments support the benefits of using two
arms and indicate that synchronous execution, in which the two arms perform together either transfers or moves, introduces only a small
overhead.  Experiments support these points and show that the
scalable method can quickly compute solutions close to the optimal for
the considered setup.

\end{abstract}

%\kiril{I've temporarily changed the format to make the display in overleaf slightly more convenient.}

\blfootnote{Work by Rahul Shome, Jingjin Yu and Kostas Bekris has been supported in part by NSF CNS-1330789, IIS-1617744 and IIS-1734419. Work by Dan Halperin has been supported in part by the Israel Science Foundation grant~825/15, the Blavatnik Computer Science Research Fund, the Blavatnik Interdisciplinary Cyber Research Center at Tel Aviv University, Yandex and Facebook.}

\vspace{-.5in}

\section{Introduction}
\label{sec:intro}
%Advances in robotics have been pushing the boundaries of the
%application of robots to automation. There has been a lot of interest
%in robotic picking in this context\cite{correll2016analysis} and
%significant recent efforts have been made for facing the challenges of
%manipulation and grasping.

\begin{figure}[t]\vspace{-0.15in}
	\centering
	\includegraphics[width=0.8\textwidth]{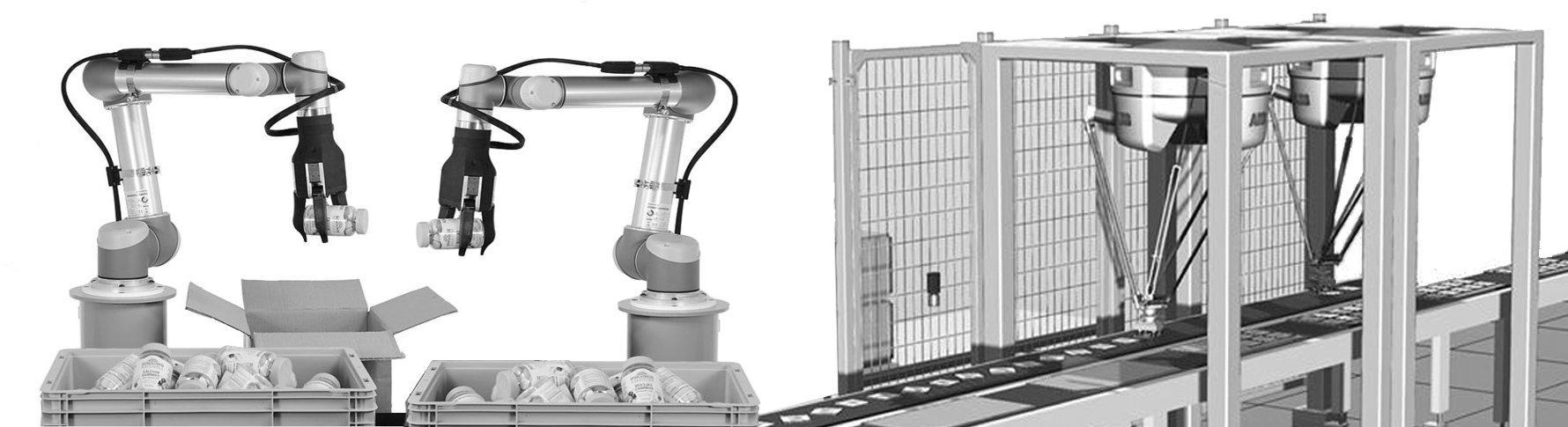}\vspace{-0.1in}
	\caption{Example of dual-arm setups that can utilize
	algorithms proposed in this work.}
	\label{fig:two_arms}\vspace{-0.25in}
\end{figure}

Automation tasks in industrial and service robotics, such as product packing or sorting, often require sets of objects to be arranged in
specific poses on a planar surface. Efficient and high-quality
single-arm solutions have been proposed for such
setups \cite{193}. The proliferation of robot arms, however, including dual-arm setups, implies that industrial settings can utilize multiple robots in the same workspace (Fig \ref{fig:two_arms}). This work explores a) the benefits of coordinated dual-arm rearrangement versus single-arm, b) the combinatorial challenges involved and c) computationally efficient, high-quality and scalable methods.

%While the benefits seem obvious, even for two arms, the challenge
%lies in examining the combinatorial sub-structure of this integrated
%task and motion planning problem.
 
%The current work focuses on rearrangement problems involving two
%arms.  Compared to the sort of solutions that yield high quality
%single arm solutions to these problems, it will be demonstrated that
%significant solution quality improvements can be made, which
%motivates the need to find effective algorithms that can solve the
%dual-arm rearrangement problem.  We seek to bound the cost of $k$-arm
%manipulation.  Note that $c_t$ may be used to capture distance
%(energy) cost or time cost.

%Due to lying at the confluence of various challenging problem
%domains, namely multi-robot motion planning, object rearrangement
%task planning, and combinatorial search, the problem at hand demands
%closer inspection through these lenses. It is desirable to obtain
%some guarantees about the benefits of using more than one arm, before
%endeavoring to study the problem.

A motivating point is that the coordinated use of multiple arms can
result in significant improvements in efficiency. This arises from the following argument.

\vspace{-.1in}
\begin{lemma}\label{l:k-arm-lower-bound}
There are classes of tabletop rearrangement problems, where a $k$-arm
($k \ge 2$) solution can be arbitrarily better than the optimal
single-arm one.
\vspace{-.3in}
\end{lemma}

For instance, assume two arms that have full (overhand) access to a 
unit square planar tabletop. There are $n$ objects on the table, 
divided into two groups of $\frac{n}{2}$ each. Objects in each group 
are $\varepsilon$-close to each other and to their goals. Let the 
distance between the two groups be on the order of $1$, i.e., the two 
groups are at opposite ends of the table. The initial position of each 
end-effector is $\varepsilon$-close to a group of objects. Let the cost 
of each pick and drop action be $c_{pd}$, while moving the 
end-effector costs $c_t$ per unit distance. Then, the $2$-arm 
cost is no more than $2nc_{pd} + 2n\varepsilon c_t $.  A single arm 
solution costs at least $2nc_{pd} + (2n-1)\varepsilon c_t + 
c_t$. If $c_{pd}$ and $\varepsilon$ are sufficiently small, the $2$-arm 
cost can be arbitrarily better than the single-arm one. The argument 
also extends to $k$-arms relative to $(k-1)$-arms.

In most practical setups, the expectation is that a $2$-arm solution
will be close to half the cost (e.g., time-wise) of the single-arm 
case, which is a desirable improvement. While there 
is coordination overhead, the best $2$-arm solution cannot do worse; simply 
let one of the arms carry out the best single arm solution while the 
other remains outside the workspace. More generally:

\vspace{-.1in}
\begin{lemma}\label{l:2-arm-no-worse}
For any rearrangement problem, the best $k$-arm ($k \ge 2$) solution
cannot be worse than an optimal single arm solution.
\vspace{-.1in}
\end{lemma}

The above points motivate the development of scalable algorithmic
tools for such dual-arm rearrangement instances. This work considers
certain relaxations to achieve this objective. In particular,
monotone tabletop instances are considered, where the start and goal
object poses do not overlap.  Furthermore, the focus is on
synchronized execution of pick-and-place actions by the two arms. i.e., where the two arms simultaneously transfer (different) objects, or simultaneously move towards picking the next objects.
Theoretical arguments and experimental evaluation indicate that the
synchronization assumption does not significantly degrade solution
quality.

The first contribution is the study of the combinatorial structure of
synchronous, monotone dual-arm rearrangement. Then, a mixed integer linear programming (\milp) model is
proposed that achieves optimal coordinated solutions in this domain.
An efficient algorithm {\tt Tour\_Over\_Matching} (\algo) is proposed
that significantly improves in scalability. \algo\ first optimizes the
cost of object transfers and assigns objects to the two arms by
solving an optimal matching problem. It minimizes the cost of
moves from an object's goal to another object's start pose per
arm as a secondary objective by employing a {\tt TSP} solution. Most
of the computation time is spent on the many calls to a
lower-level motion planner that coordinates the two arms. A lazy
evaluation strategy is proposed, where candidate solution sequences
are developed first and then motion plans are evaluated for them. The
strategy results in significant computational improvement and
minimizes the calls to the coordination motion planner.  An analysis
is also performed on the expected improvement in solution quality
versus the single arm case, as well as the expected cost overhead from a
synchronous solution.

Finally, experiments for i) a simple planar picker setting, and ii) a
model of two 7-DOF Kuka arms, demonstrate a nearly two-fold
improvement against the single arm case for the proposed approach in practice. The
solutions are close to optimal for the dual-arm case and the algorithm
exhibits good scalability. The computational costs of both the optimal
and the proposed method are significantly improved from the lazy
evaluation strategy. 
% When the synchronous solutions are smoothed to
% asynchronous ones, there is not significance difference in overall
% path quality. \kiril{This statement is unclear. What is the benefit from performing this step?}

\section{Related Work}
\label{sec:background}
%Related efforts:
%
%\begin{itemize}
%\item multi-robot planning and coordination
%\item manipulation task planning in general
%\item rearrangement 
%\item combinatorial literature
%\end{itemize}
The current work dealing with dual-arm object rearrangement touches upon the challenging intersection of a variety of rich bodies of prior work. It is closely related to multi-robot planning and coordination. The challenge with multi-body planning is the high dimensionality of the configuration space. %Early work has suggested to slightly reduce the~\cite{aronov1999motion} tried dimensionality reduction strategies. 
Optimal strategies were developed for simpler instances of the problem \cite{solovey2015motion}, although in general the problem is known to be computationally hard~\cite{solovey2016hardness}. Decentralized approaches~\cite{van2005prioritized} also used velocity tuning~\cite{leroy1999multiple} to deal with these difficult instances. General multi-robot planning tries to plan for multiple high-dimensional platforms \cite{wagner2012probabilistic,Gharbi:2009fu} using sampling-based techniques. 
Recent advances provide scalable~\cite{SoloveySH16:ijrr} and asymptotically optimal~\cite{Dobson:2017aa} sampling based frameworks. 
%Assembly planning \cite{halperin2000general,sundaram2001disassembly} is another related problem dealing with multi-body planning.

%The discrete version of the multi-body problem, also known as the "pebble motion problem" has seen a lot of work \cite{kornhauser1984coordinating,auletta1999linear,goraly2010multi}. For such discrete problems on a pebble graph, feasibility times are linear and solution times are polynomial. Optimality is still challenging even in these setups. Heuristics have been used to provide high quality solutions \cite{wagner2012probabilistic,sharon2015conflict}. 

In some cases, by restricting the input of the problem to a certain type, it is possible to cast known hard instances of a problem as related algorithmic problems which have efficient solvers. For instance, unlabeled multi-robot motion planning can be reduced to pebble motion on graphs~\cite{abhs-unlabeled14}; pebble motion can be reduced to network flow~\cite{yu2016optimal}; and single-arm object rearrangement can be cast as a traveling salesman problem~\cite{193}. These provide the inspiration to closely inspect the structure of the problem to derive efficient solutions.

%\subsection{Combinatorial Literature}
%Our problem can be formulated in a variety of related classes of problems in combinatorial literature. Most of the approaches assume that all the edge weights are known, ie. the motion planning for all the edges has to be performed before any operation on the graph that uses the weights.

%\textbf{The Traveling Salesman Problem}:
In this work we leverage a connection between dual-arm rearrangement and two combinatorial problems: (1) optimal matching~\cite{edmonds1965maximum} and (2) TSP. On the surface the problem seems closely related to multi-agent TSP.
%This can itself be tweaked with assumptions on same or different start and goal positions, symmetric or asymmetric metrics, directed or undirected graphs or posing the problem as an optimization
A seminal paper~\cite{frederickson1976approximation} provides the formulation for k-TSP, with solutions that split a single tour.
Prior work \cite{rathinam2006matroid} has formulated the problem of multi-start to multi-goal k-TSP as an optimization task. Some work~\cite{friggstad2013multiple} deals with asymmetric edge weights which are more relevant to the problems of our interest.
 
%\textbf{Errand Scheduling}:
%If the object transfer is seen as an errand \cite{slavik1997errand} then a graph constructed with nodes being object-arm assignments and edges implying the order of coordinated execution. 
%There has been prior work \cite{garg2000polylogarithmic} that operates over the assignment graph(where pairwise assignments are nodes), an object's group consists of sets of nodes that include an object

The multi-arm rearrangement can also be posed as an instance of multi vehicle pickup and delivery (PDP)~\cite{parragh2008survey}. 
%Most of the work has been motivated by ride-sharing applications and timed task scheduling, so there are a lot of variations that deal with time windows but this line of research provides some ILP formulations for the problem.
Prior work~\cite{coltin2014multi} has applied the PDP problem to robots, taking into account time windows and robot-robot transfers.
Some seminal work~\cite{lenstra1981complexity,savelsbergh1995general} has also explored its complexity, and concedes to the hardness of the problem, while others have studied cost bounds~\cite{TrePavFra13}. ILP formulations~\cite{savelsbergh1995general} have also been proposed. 
%The existing work that delves into the combinatorial challenges of the problem, do not account for the costs incurred due to coordination of the agents during task execution. 
%Pickup and Delivery problems under track contention \cite{caricato2003parallel}, makes an attempt to reason about the coordination of the robots over solutions obtained with Tabu Search.
Typically this line of work ignores coordination costs, though some methods~\cite{caricato2003parallel} reason about it once candidate solutions are obtained.

% There has been a a lot of work in motion planning for manipulators and movable objects. 
Navigation among movable objects deals with the combinatorial challenges of multiple objects~\cite{wilfong1991motion,van2009path} and has been shown to be a hard problem, and extended to manipulation applications~\cite{stilman2007manipulation}. Despite a lot of interesting work on challenges of manipulation and grasp planning, the current work shall make assumptions that avoid complexities arising from them. Manipulators opened the applications of rearrangement planning~\cite{ben1998practical,ota2004rearrangement}, including instances where objects can be grasped only once or monotone~\cite{stilman2007manipulation}, as well as non-monotone instances~\cite{havur2014geometric,srivastava2014combined}. Efficient solutions to assembly planning problems~\cite{Wilson:1994fk,Halperin:2000uq} typically assumes monotonicity, as without it the problem becomes much more difficult. Recent work has dealt with the hard instances of task planning~\cite{berenson2011task,cohen2014single} and rearrangement planning~\cite{krontiris2015dealing,krontiris2016efficiently,193}. Sampling-based task planning has made a recent push towards guarantees of optimality~\cite{vega2016asymptotically,schmitt2017optimal}. These are broader approaches that are invariant to the combinatorial structure of the application domain. The current work draws inspiration from these varied lines of research.

\commentadd{
General task planning methods are unaware of the underlying structure studied in this work. Single-arm rearrangement solutions will also not effective in this setting. The current work tries to bridge this gap and provide insights regarding the structure of dual-arm rearrangement. Under assumptions that enable this study, an efficient solution emerges for this problem.
}

\section{Problem Setup and Notation}
\label{sec:problem}
Consider a planar surface and a set of $n$ rigid objects $ \objectset=\{ \object_1, \object_2 \cdots \object_n\}$  that can rest on the surface in stable poses $p_i \in \Pspace_i \subset SE(3)$. The arrangement space $ \Aspace = \Pspace_1 \times \Pspace_2 \ldots \times \Pspace_n $ is the Cartesian product of all $\Pspace_i$, where $ \arrangement = (\pose_1, \ldots, \pose_n) \in \Aspace $. In valid arrangements $ \Afree \subset \Aspace$, the objects are pairwise disjoint.

\begin{wrapfigure}{r}{2.4in}
    \vspace{-.3in}
	\centering
	\includegraphics[width=1.115in,trim={10cm 3cm 8cm 12cm},clip]{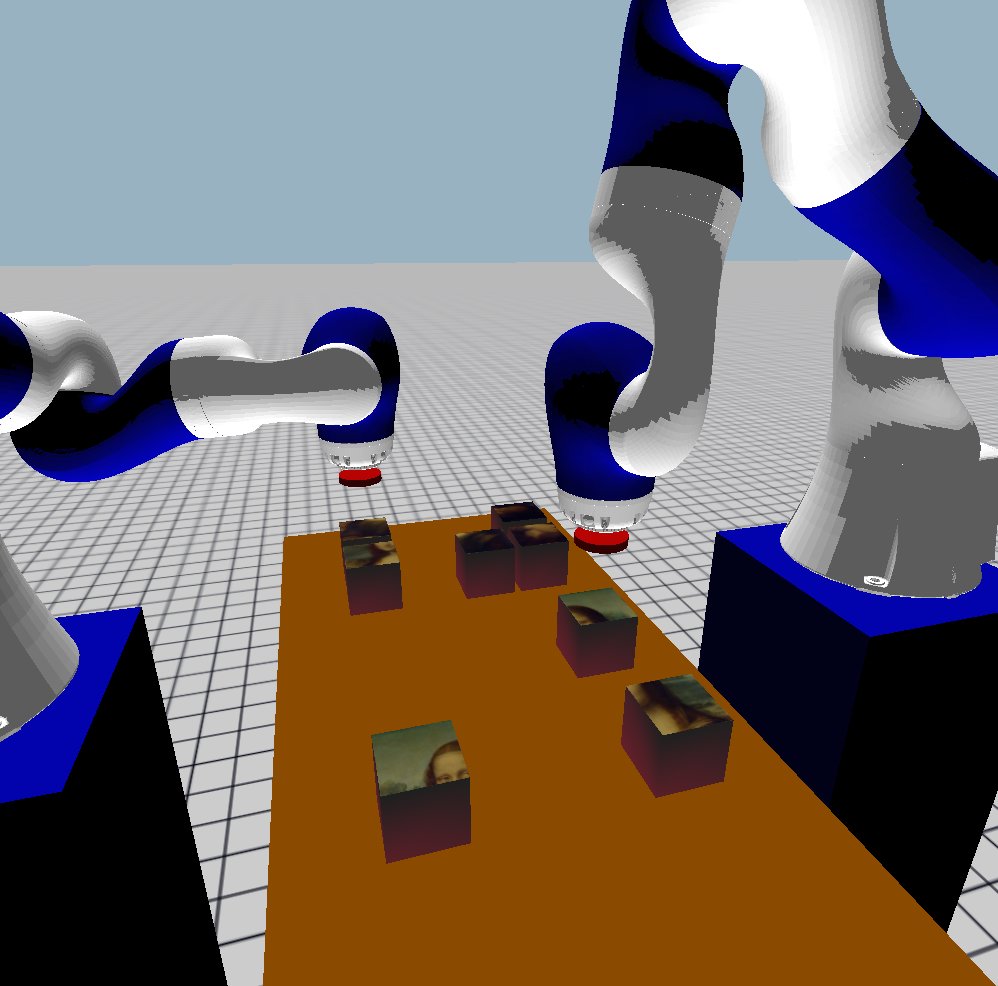}
	\includegraphics[width=1.115in,trim={10cm 3cm 8cm 12cm},clip]{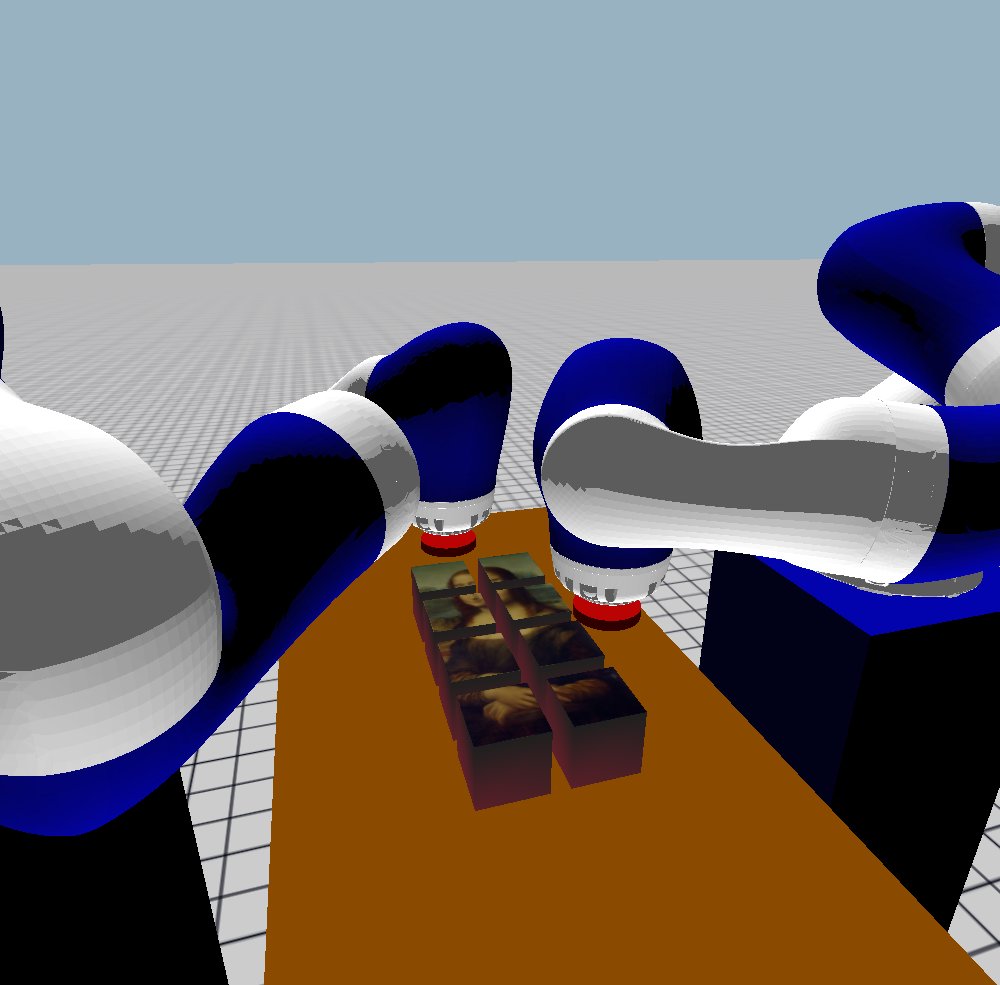}
    \vspace{-.15in}
	\caption{An example of object rearrangement involving two robotic arms. Initial (left) and final (right) object configuration.}
	\label{fig:rearrangement}
    \vspace{-.35in}
\end{wrapfigure}
Two robot arms $\arm^1$ and $\arm^2$ can 
% pick, raise above the surface and place the objects. 
pick and place the objects from and to the surface.
% \kiril{The last statement is unclear, particularly the purpose of the word "raise". Can it be rewritten into "...can pick and place objects from and to the surface."?}  
% THE OLD SENTENCE$\cfree^k$ is the set of collision-free configurations for arm $ \arm^k $, where $ q^k \in \cfree^k %\subset \reals^d $ (while ignoring possible collisions with  objects or the other arm).
$\cfree^k$ is the set of collision-free configurations for arm $\arm^k$ (while ignoring possible collisions with  objects or the other arm), and is assumed here to be a subset of $\reals^d $.
A path for $\arm^k$ is denoted as $ \pi^k : [0,1] \rightarrow \cfree^k$ and includes picking and placing actions.  Let $ \cfree^i(\pi^j) $ represent the collision free $\mathbb{C}$-space for $\arm^i$, given that $\arm^j$ moves along $\pi^j $. This is a function of the paths' parametrization. The space of dual-arm paths $\mathcal{D}$ denotes pairs of paths for the two arms: $D = (\pi^1,\pi^2) \in \mathcal{D}$. Then, $\arrangement(\ainit, D)$ is the resulting arrangement when the objects are at $\ainit$ and moved according to $D$.   Let $ \cost(D) : \mathcal{D} \rightarrow \reals $ be a cost function over dual-arm paths. 

\vspace{-.1in}
{\definition [Optimal Dual-arm Rearrangement] Given arms $\arm^1$, $\arm^2$ and objects $\objectset$ to be moved from initial arrangement $ \ainit \in \Afree$ to target arrangement $ \atarget \in \Afree $, the optimal dual-arm rearrangement problem asks for a dual-arm path $D^* \in \mathcal{D}$, which satisfies the following constraint:

\vspace{-.15in} 
\begin{equation}
D^* =  (\pi^{*1}, \pi^{*2}) \ |\ \arrangement(\ainit, D^*) = \atarget \textrm{ and } \pi^{*i} \in \cfree^i(\pi^{*j}) \vspace{-.05in}
\label{eq:dualarm_constraint}
\end{equation}
\noindent and optimizes a cost function: $D^* = \underset{ \forall D \in \mathcal{D}}{ \mathtt{argmin} } \ \cost(D).$}

A set of assumptions are introduced to deal with the problem's combinatorics. The reachable task-space $\taskspace^k \subset SE(3)$ of arm $\arm^k$ is the set of $SE(3)$ poses that objects attached to the arm's end effector can acquire.

% \vspace{-.15in}
\commentdel{
{\textbf{Assumption} [Reachability] All objects are reachable by both arms at $\ainit$ and $\atarget$: $\forall\ \pose_i \in \ainit \textrm{ and } \forall\ \pose_j \in \atarget \textrm{ and } \forall\ k \in [1,2]: \pose_i, \pose_j \subset \taskspace^k $.}
}

Let the ordered set of objects moved during the arm path $ \pi^i $ be denoted as $ \moveset(\pi^k) $. In general, an object can appear many times in $ \moveset(\pi^k) $. The current work, however, focuses on monotone instances, where each object is moved once.

\vspace{-.15in}
{\assumption [Monotonicity] There are dual-arm paths $D = (\pi^1,\pi^2)$ that satisfy Eq. \ref{eq:dualarm_constraint}, where each object $ \object_i \in \objectset $ appears once in $\moveset(\pi^1)$ or $ \moveset(\pi^2)$.}

\commentadd{
For the problem to be solvable, all objects are reachable by at least one arm at both $\ainit$ and $\atarget$: $\forall\ \pose_i \in \ainit \textrm{ and } \forall\ \pose_j \in \atarget \textrm{ and } \exists\ k \in [1,2]: \pose_i, \pose_j \subset \taskspace^k $.
}
The focus will be on simultaneous execution of transfer and move paths. 

\noindent\textbf{Transfers:} Dual-arm paths $T(\pi^1_i,\pi^2_i) \in \mathcal{D}$, where $ \moveset(\pi^k_i) =  \object^k_i $ and each $m^k$:
\begin{myitem}
\item[$-$]\ \ starts the path in contact with an object $\object^k_i$ at its initial pose in $\ainit$,
\item[$-$]\ \ and completes it in contact with object $ \object^k_i $ at its final pose in $\atarget$. 
\end{myitem}

\noindent\textbf{Moves or Transits:} Paths $M(\pi^1_{i\rightarrow i^{\prime}},\pi^2_{i\rightarrow i^{\prime}}) \in \mathcal{D}$, $ \moveset(\pi^1_{i\rightarrow i^{\prime}}) = \emptyset $, and each $\arm^k$:
\begin{myitem}
\item[$-$]\ \  starts in contact with object $\object^k_i$ at its final pose in $\atarget$,
\item[$-$]\ \  and completes it in contact with object $\object^k_{i^{\prime}}$ at its initial pose in $\ainit$.
\end{myitem}

\vspace{-.1in}
{\assumption [Synchronicity] Consider dual-arm paths, which can be decomposed into a sequence of simultaneous transfers and moves for both arms:
\vspace{-.1in}
\begin{equation}
D=\left(T(\pi^1_1,\pi^2_1),M(\pi^1_{1\rightarrow 2},\pi^2_{1\rightarrow 2}), \ldots , M(\pi^1_{\ell-1\rightarrow \ell},\pi^2_{\ell-1\rightarrow \ell}), T(\pi^1_\ell,\pi^2_\ell)\right).\vspace{-.05in}
\label{eq:synchronicity}
\end{equation}}
\noindent For simplicity, Eq.~\ref{eq:synchronicity} does not include an initial move from $q^k_{\rm safe} \in \cfree^k$ and a final move back to $q^k_{\rm safe}$. An odd number of objects can also be easily handled. Then, the sequence of object pairs moved during a dual-arm path as in Eq. \ref{eq:synchronicity} is: \vspace{-.1in} 
% $$ \scoma(D) = \left(  \coma_i = (\object_i^1,\object_i^2) \ |\ i,j\in [1,\cdots,\ell], \bigcup_i (\object_i^1 \cup \object_i^2) = \objectset, \object_i^1 \cap \object_j^1 = \object_i^2 \cap \object_j^2 = \emptyset \right). \vspace{-.15in}$$
$$ \scoma(D) = \left(  \coma_i = (\object_i^1,\object_i^2) \ |\ i,j\in [1\cdots \ell], \bigcup_i (\object_i^1 \cup \object_i^2) = \objectset, \forall k,k'\in[1,2], \object_i^k \neq \object_j^{k'} \right). \vspace{-.1in}$$

\noindent Given the pairs of objects $\coma_i$, it is possible to express a transfer as $T(\coma_i)$ and a move as $M(\coma_{i\rightarrow j})$.  Then, $D(\scoma)$ is the synchronous, monotone dual-arm path due to $\scoma = (\coma_1, \ldots, \coma_{\ell})$, i.e., $D(\scoma) = (T(\coma_1),M(\coma_{1\rightarrow 2}),\ldots,M(\coma_{\ell-1\rightarrow \ell}),T(\coma_\ell)).$

\vspace{-.1in}
{\assumption [Object Non-Interactivity] There are collision-free transfers $T(\coma_i)$ and moves $M(\coma_{i\rightarrow i^{\prime}})$ regardless of the object poses in $\ainit$ and $\atarget$.}
\commentadd{This entails that there is no interaction between the $n$ resting objects and the arms during transits. Similarly, there are no interactions between the arm-object system and the $n-2$ resting objects during the transfers. Collisions involving the arms, static obstacles and picked objects are always considered.}
\commentdel{
This assumption is valid in tabletop setups where the arms can raise the picked objects above the resting surface. 
% Furthermore, this assumption implies that transfers and moves involving object poses in $\ainit, \atarget$ are always feasible and the challenge is deciding the appropriate sequence of object-to-arm assignments.
This assumption lets the study focus on the challenge of deciding the appropriate sequence of object-to-arm assignments.
}

The metric this work focuses on relates to makespan and minimizes the sum of the longest distances traveled by the arms in each synchronized operation. Let $ \| \pi^k \| $ denote the Euclidean arc length in $\cfree^k  \subset \reals^d$ of path $ \pi^k $. Then, for transfers $\cost( T(\pi^1_i,\pi^2_i) ) = \max\{ \|\pi^1_i\|, \|\pi^2_i\| \}$. Similarly, $\cost( M(\pi^1_{i\rightarrow i^{\prime}},\pi^2_{i\rightarrow i^{\prime}}) ) = \max\{ \|\pi^1_{i\rightarrow i'}\|, \|\pi^2_{i\rightarrow i'}\| \}$. Then, over the entire dual-arm path $ \D $ define:
\vspace{-.15in}
\begin{equation}
\cost(\D) 
% = \cost_{ T} + \cost_{ M} 
= \sum_{i=1}^{\ell}   \cost(T(\coma_i))   + \sum_{i=1}^{\ell-1}   \cost(M(\coma_{i\rightarrow i+1})). 
\vspace{-.1in}
\label{eq:cost_function}
\end{equation}
Note that the \textit{transfer} costs do not depend on the order with which the objects are moved but only on the assignment of objects to arms. The \textit{transit} costs arise out of the specific ordering in $ \scoma(D)$. 
Then, for the setup of Definition 1 and under Assumptions 1-3, the problem is to compute the optimal sequence of object pairs $\scoma^*$ so that $D(\scoma^*)$ satisfies Eq.~\ref{eq:dualarm_constraint} and minimizes the cost of Eq.~\ref{eq:cost_function}.

\commentdel{Assumptions 2 and 3 will be partly relaxed in Section \ref{sec:integration}. Assumption 1 is also effectively relaxed to cases where at least one arm can transfer every object.}
% , given a smoothing process for achieving an asynchronous solution and a lazy motion planning process for computing transfers and moves that can be informed of object poses. 

\commentadd{
\noindent \textbf{Note on Assumptions:} This work restricts the study to a class of monotone problems that relate to a range of industrial packing and stowing applications. The monotonicity assumption is often used in manageable variants of well-studied problems~\cite{Wilson:1994fk,Halperin:2000uq}. A monotone solution also implies that every object's start and target is reachable by at least one arm. To deal with cases where the number of reachable objects is unbalanced between the two arms, a {\tt NO\_ACT} task assignment is introduced and considered in Section 6. 

The synchronicity assumption allows to study the combinatorial challenges of the problem, which do not relate to the choice of time synchronization of different picks and placements. Section \ref{sec:integration} describes the use of $\drrtstar$\cite{Dobson:2017aa} as the underlying motion planner that can discover solutions that can synchronize arm motions for simultaneous picks, and simultaneous placements. The synchronicity assumption is relaxed through smoothing (Section \ref{sec:integration}).

The non-interactivity assumption comes up naturally in planar tabletop scenarios with top-down picks or delta robots. Such scenarios are popular in industrial settings. Once the object is raised from the resting surface, transporting it to its target does not introduce interactions with the other resting objects. This assumption is also relaxed in Section \ref{sec:integration}, with a lazy variant of the proposed method. Once a complete candidate solution is obtained, collision checking can be done with everything in the scene.

Overall, under the assumptions, the current work identifies a problem structure, which allows arguments pertaining to the search space, completeness, and optimality. Nevertheless, the smoothed, lazy variant of the proposed method will still return effective solutions in practice, even if these assumptions do not hold.

% The validation of the final candidate solution enforces that the algorithm keeps trying till a collision-free solution is found. The assumptions enable the current work to study in detail the problem, which is otherwise mostly intractable. 

}

\section{Baseline Approaches and Size of Search Space}
\label{sec:baseline}
This section highlights two optimal strategies to discover $\D(\scoma^*)$: a) exhaustive search, which reveals the search space of all possible sequences of object pairs $ \scoma $ and b) an MILP model. Both alternatives, however, suffer from scalability issues as for each possible assignment, it is necessary to solve a coordinated motion planning problem for the arms. This motivates minimizing the number of assignments $\coma$ considered, and the number of motion planning queries it requires for discovering $\D(\scoma^*)$, while still aiming for high quality solutions.

\begin{wrapfigure}{r}{2.4in}
    \vspace{-.1in}
	\centering
	\includegraphics[width=2.3in]{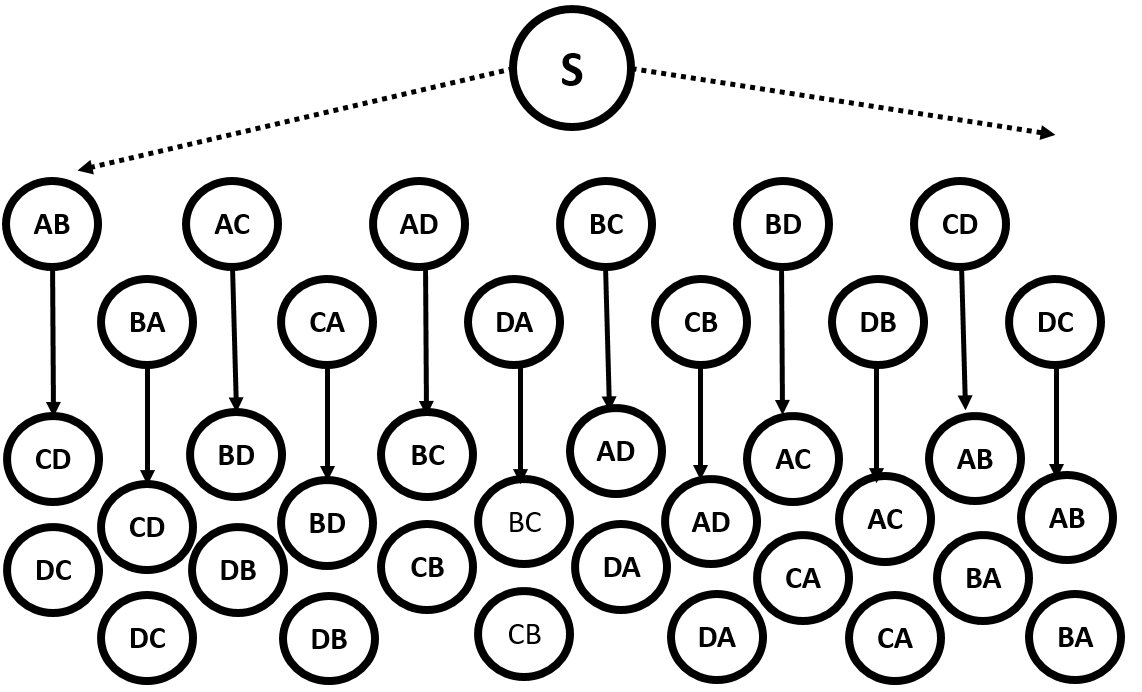}
    \vspace{-.15in}
	\caption{Search tree for 4 objects. In node $XY$, $m^1$ transfers object $X$, and $m^2$ transfers $Y$. An edge is a transit to the next node. }
	\label{fig:backtracking}
    \vspace{-.3in}
\end{wrapfigure}

\noindent\textbf{Exhaustive Search}: The exhaustive search approach shown in Fig. \ref{fig:backtracking} is a brute force expansion of all possible sequences of object pairs $ \scoma $. Nodes correspond to transfers $T(\coma_i)$ and edges are moves $M(\coma_{i\rightarrow i{^\prime}})$. The approach evaluates the cost for all possible branches to return the best sequence $\scoma$. The total number of nodes is in the order of $O(n!)$, expressed over the levels $L$ of the search tree as: 
% \kiril{It may be beneficial to explain the formula below, or perhaps move it to the appendix to save space and avoid explaining it here.}:

\vspace{-0.25in}
\begin{equation*}
{\permu[n]{2}} + {\permu[n]{2}}\times{\permu[n-2]{2}} + ... + {\permu[n]{2}}\times{\permu[n-2]{2}}\times{\permu[n-4]{2}}...\times{\permu[2]{2}} =\sum_{L=1}^{n/2} {\ \prod_{k=0}^{2L-1}{(n-k)}},
\vspace{-0.15in}
\end{equation*}
where $\permu[n]{k}$ denotes the number of $k$-permutations of $n$.  

Motion plans can be reused, however, and repeated occurrences of $T(\coma_i)$ and $M(\coma_{i\rightarrow i^{\prime}})$ should be counted only once for a total of: 
\begin{myitem}
\item[$-$] $  {\permu[n]{2}} $ \textit{transfers} of objects, and 
\item[$-$] $ {{\permu[n]{2}}\times\permu[n-2]{2}} $ \textit{transits} between all possible valid ordered pairs of $ \coma $.
\end{myitem}
Additional motion plans are needed for the initial move from $q_{\rm safe}^k$ and the return to it at the end of the process, introducing $ 2\times\permu[n]{2} $ transits:

\vspace{-0.35in}
\begin{equation}
\textrm{\# of Transfers} + \textrm{\# of Moves} =  {\permu[n]{2}} +   \Big{(}({{\permu[n]{2}}\times\permu[n-2]{2}}) + (2\times{\permu[n]{2}})\Big{)}
\label{eq:mpcount}
\vspace{-0.2in}
\end{equation}

This returns optimal synchronized solution but performs an exhaustive search and requires exponentially many calls to a motion planner.

\noindent\textbf{MILP Formulation}:
Mixed Integer Linear Programming (\milp) formulations can utilize highly optimized solvers \cite{gurobi}. Prior work has applied these techniques for solving m-TSP \cite{rathinam2006matroid,friggstad2013multiple} and pickup-and-delivery problems \cite{coltin2014multi,savelsbergh1995general}, but viewed these problems in a decoupled manner. This work outlines an \milp formulation for the synchronized dual-arm rearrangement problem that reasons about coordination costs arising from arm interactions in a shared workspace.

\textit{Graph Representation:}
The problem can be represented as a directed graph where each vertex $v = \coma_v = (\object_v^1, \object_v^2)$ corresponds to a transfer $ T(\coma_u) $ and edges $e(u,v)$ are valid moves $M(\coma_{u\rightarrow v})$. A valid edge $e(u,v)$ is one where an object does not appear more than once in the transfers of nodes $u$ and $v$.  The cost of a directed edge $e(u,v)$ encodes both the cost of the transfer  $ T(\coma_v) $ and the cost of the move $M(\coma_{u\rightarrow v})$. There is also a vertex $ \startq $, which connects moves from and to the safe arm configurations $q^k_{\rm safe}$.  The directed graph $ \hat{G} (\hat{V},\hat{E}) $ is defined:

\vspace{-.3in}
\begin{align*}
\hspace{0.5in} \hat{V} = \{ v = \coma_v = (\object_v^1,\object_v^2)\ | \  \forall\ \object_v^1,\object_v^2 \in \objectset, \object_v^1 \neq \object_v^2 \} \cup \{S\}\\
\hat{E} = \{  e(u,v)\ | \ \forall\ u,v\in \hat{V} \textrm{ so that } u\neq v,\ \object_v^k \neq \object_u^{\ell}\ \forall\ k, \ell \in [1,2] \}\\
\cup\ \{e(\startq,v)\ \forall\ v\in\hat{V}\setminus \startq\} \cup \{e(v,\startq)\ \forall\ v\in\hat{V}\setminus \startq\}\\
cost_{e(u,v)} = \cost(u) + \cost(u,v) =\cost(T(\coma_u)) + \cost(M(\coma_{u\rightarrow v}))
\end{align*}

\vspace{-.2in}
Let $\cost(\startq) = 0$. The total number of motion planning queries needed to be answered to define the edge costs is expressed in Eq.~\ref{eq:mpcount}. The formulation proposed in this section tries to ensure the discovery of $ \scoma^* $ on $ \hat{G} $ as a tour that starts and ends at $ \startq $, while traversing each vertex corresponding to $ \scoma^* $. To provide the {\tt MILP} formulation, define $ \delta_{\rm in}(v) $ as the in-edge set $v$, and  $ \delta_{\rm out}(v) $ as the out-edge set. Then, $ \gamma(\object) $ is the object coverage set $ \gamma(\object) = \{ e(u,v) \ |\ e\in\hat{E}, \object\in  \coma_u  \} $, i.e., all the edges that transfer $ \object $.

\textit{Model:} Set the optimization objective as: $\ \ \min \sum_{e \in \hat{E}} cost_e x_e\  $\hspace{.81in} [A]\\ 
Eq. [B] below defines indicator variables. Eqs. [C-E] ensure edge-flow conserved tours. Eqs. [F-G] force $\startq$ to be part of the tour. Eq. [H] transfers every object only once. Eq. [I] lazily enforces the tour to be of length $ \frac{n}{2} + 1 $. While the number of motion-planning queries to be solved is the same as in exhaustive search, efficient {\milp} solvers \cite{gurobi} provide a more scalable search process.

\vspace{-.25in}
\noindent\begin{minipage}{.5\textwidth}
\begin{align*}
x_e \in \{ 0,1 \}& \ \ \forall e \in \hat{E} \tag*{[B]}\\
\sum_{e\in\delta_{\rm in}(v)} x_e \leq 1& \ \ \forall v\in \hat{V} \tag*{[C]}\\
\sum_{e\in\delta_{\rm out}(v)} x_e \leq 1& \ \ \forall v\in \hat{V} \tag*{[D]}\\
\sum_{e\in\delta_{\rm in}(v)}x_e = \sum_{e\in\delta_{\rm out}(v)}x_e& \ \ \forall v\in \hat{V} \tag*{[E]}\\
\end{align*}
\end{minipage}
\noindent\begin{minipage}{.5\textwidth}
\begin{align*}
\sum_{e\in\delta_{\rm in}(\startq)} x_e = 1& \tag*{[F]}\\
\sum_{e\in\delta_{\rm out}(\startq)} x_e = 1& \tag*{[G]}\\
\sum_{e\in\gamma(\object)} x_e = 1& \ \  \forall \object\in\objectset \tag*{[H]}\\
\sum_{e(u,v)\in \mathfrak{T}} x_e < |\mathfrak{T}| & \ \  \forall \mathfrak{T} \subset \hat{V}, |\mathfrak{T}| \leq \frac{n}{2} \tag*{[I]}
\label{model:milp}
\end{align*}
\end{minipage}

\section{Efficient Solution via Tour over Matching}
\label{sec:approximation}

The optimal baseline methods described above highlight the problem's complexity. Both methods suffer from the large number of motion-planning queries they have to perform to compute the cost measures on the corresponding search structures. For this purpose it needs to be seen whether it is possible to decompose the problem into solvable sub-problems. 
%%%%%%%%%%%%%%%%%%%%%%%%%%%%%%%%%%%%%%%%%%%%%%%%%%%%%%%%%%%%%%
%%%%%%%%%%%%%%%%%%%%%%%%%%%%%%%%%%%%%%%%%%%%%%%%%%%%%%%%%%%%%%
%%%%%%%%%%%%%%%%%%%%%%%%%%%%%%%%%%%%%%%%%%%%%%%%%%%%%%%%%%%%%%
%%%%%%%%%%%%%%%%%%%%%%%%%%%%%%%%%%%%%%%%%%%%%%%%%%%%%%%%%%%%%%

\noindent\textbf{Importance of Transfers}: In order to draw some insight, consider again the tabletop setup with a general cost measure of $c_t$ per unit distance.
Lemma~\ref{l:k-arm-lower-bound} suggests that under certain conditions, there 
may not be a meaningful bound on the performance ratio between a $k$-arm 
solution and a single-arm solution. This motivates the examination of another
often used setting---randomly chosen non-overlapping 
start and goal locations for $n$ objects (within a unit square). In order to derive a 
meaningful bound on the benefit of using a $2$-arm solution 
to a single-arm solution, we first derive a conservative cost of a single-arm solution. 
A single-arm optimal cost has three main parts: 1) the portion of the transfer cost involving the pickup and 
drop-off of the $n$ objects with a cost of $C_{pd} = nc_{pd}$, 2) the remaining transfer cost from 
start to goal for all objects $C_{sg}$, and 3) the transit cost 
going from the goals to starts $C_{gs}$. The single arm cost is  
\vspace{-0.1in}
\begin{equation}
\label{eq:single-cost-exp}
C_{\rm single} = nc_{pd} + C_{sg} + C_{gs}.
\vspace{-0.1in}
\end{equation}
To estimate Eq.~\ref{eq:single-cost-exp}, first note that the randomized setup 
will allow us to obtain the expected $C_{sg}$\cite{santalo2004integral} as
$\frac{2 + \sqrt{2} + 5\ln(1+\sqrt{2})}{15}nc_t \approx 0.52nc_t$.
Approximate $C_{gs}$ by simply computing an optimal assignment of the 
goals to the starts of the objects excluding the matching of the same start 
and goal. Denoting the distance cost of this matching as $C_{gs}^M$, clearly
$C_{gs} > C_{gs}^M$ because the paths produced by the matching may form 
multiple closed loops instead of the desired single loop that connects all 
starts and goals. However, the number of loops produced by the matching 
procedure is on the order of $\ln n$ and therefore, $C_{gs} < C_{gs}^M + 
c_t\ln n$, by~\cite{TrePavFra13}. By \cite{AjtKomTus84}, 
$C_{gs}^M = \Theta(\sqrt{n\ln n})$. Putting these together, we have,

{\centerline
{
$\Theta(c_t\sqrt{n\ln n}) = C_{gs}^M < C_{gs} < C_{gs}^M + c_t\ln n 
= \Theta(c_t\sqrt{n\ln n}) + c_t\ln n$
}
}

\noindent which implies that $C_{gs} \approx C_{gs}^M$ because for large $n$, $\ln n
\ll \sqrt{n\ln n}$. It is estimated in \cite{yu2015target} that $C_{gs}^M 
\approx 0.44\sqrt{n\ln n}c_t$ for large $n$. Therefore, 

% \noindent Noting that the $0.44\sqrt{n\ln n}c_t$ term may also be ignored for large $n$,
% we have 
\vspace{-0.1in}
\begin{equation}\label{eq:single-cost-simple}
C_{\rm single} \approx nc_{pd} + (0.52n + 0.44\sqrt{n\ln n})c_t \approx (c_{pd} + 0.52c_t)n
% \vspace{-0.1in}
\end{equation}
noting that the $0.44\sqrt{n\ln n}c_t$ term may also be ignored for large $n$. The cost of dual arm solutions will be analyzed in Section \ref{sec:cost_bounds}.
{
\lemma For large $n$, the transfers dominate the cost of the solution.
\label{lem:transferdomination}
}
%%%%%%%%%%%%%%%%%%%%%%%%%%%%%%%%%%%%%%%%%%%%%%%%%%%%%%%%%%%%%%
%%%%%%%%%%%%%%%%%%%%%%%%%%%%%%%%%%%%%%%%%%%%%%%%%%%%%%%%%%%%%%
%%%%%%%%%%%%%%%%%%%%%%%%%%%%%%%%%%%%%%%%%%%%%%%%%%%%%%%%%%%%%%
%%%%%%%%%%%%%%%%%%%%%%%%%%%%%%%%%%%%%%%%%%%%%%%%%%%%%%%%%%%%%%

We formulate a strategy to reflect this importance of transfers.
The proposed approach gives up on the optimality of the complete problem, instead focusing on a high-quality solution, which:
\begin{myitem}
\item[$-$] first optimizes transfers and selects an assignment of object pairs to arms, 
\item[$-$] and then considers move costs and optimizes the schedule of assignments.
\end{myitem} 
This ends up scaling better by effectively reducing the size of the search space and performing fewer motion planning queries. It does so by optimizing a related cost objective and taking advantage of efficient polynomial-time algorithms. 

\begin{minipage}{0.48\textwidth}
  \begin{align}
  \label{eq:transfergraph}
  \begin{split}
  \graph(\nodes, \edges)\\
  \nodes = \{ \node=\object, \forall \object \in \objectset \}\\
  \edges = \{ \edge(u, v): \coma = (u,v), \forall u,v\in\nodes \}\\
  \cost(\edge(u,v)) = \cost(T(\coma=(u,v)))
  \end{split}
  \end{align}
\end{minipage}
\begin{minipage}{0.48\textwidth}
  \begin{align}
  \label{eq:transitgraph}
  \begin{split}
  \tspgraph(\tspnodes, \tspedges)\\
  \tspnodes = \{ \node=\coma, \forall \coma \in \scomaset^* \} \cup \{\coma_S\} \\
  \tspedges = \{ \edge(u, v): \coma_{u\rightarrow v}, \forall u,v\in\tspnodes \}\\
  \cost(\edge(u,v)) = \cost(M(\coma_{u\rightarrow v}))
  \end{split}
  \end{align}
\end{minipage}
\vspace{0.05in}

% \begin{align*}
% \ainit \rightarrow \atarget \\
% \scomaset^*\ \mathtt{on}\ \graph \\
% \scoma^+ : \tour \ \mathtt{on}\ \tspgraph
% \end{align*}

\begin{wrapfigure}{r}{2.5in}\vspace{-0.5in}
% \begin{figure}[H]
	\begin{center}
		\includegraphics[width=2.5in]{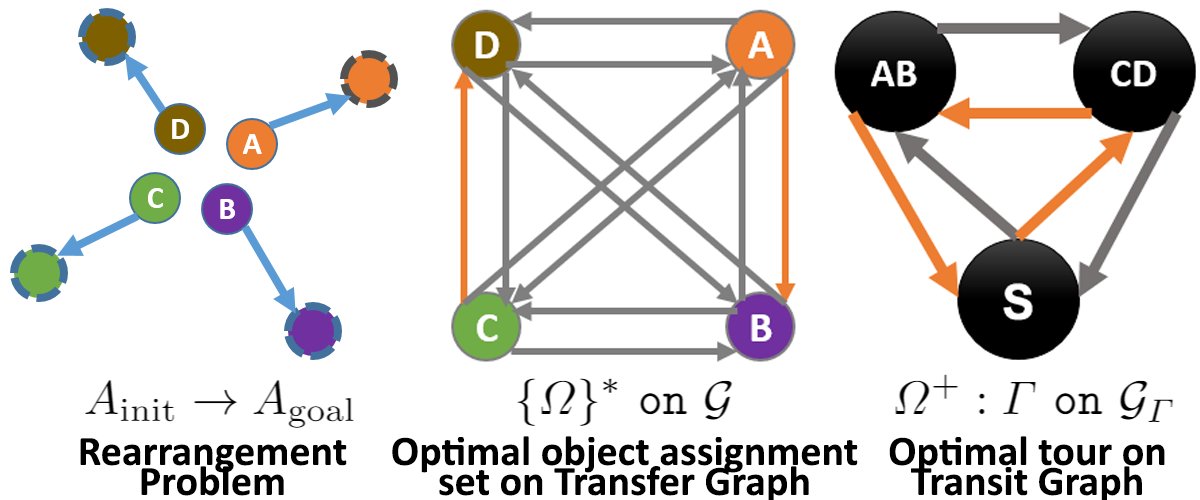}
	\end{center}\vspace{-0.2in}
	\caption{(left) A dual-arm rearrangement problem. (middle) The same problem as minimum weight edge matching on a fully connected directed graph of \textit{transfers}. (right) The ordering of the object-arm assignments from an optimal tour over a \textit{transit} graph.}\vspace{-0.2in}
	\label{fig:edge_matching}
% \end{figure}
\vspace{-0.1in}
\end{wrapfigure}

\noindent\textbf{Foundations:} Consider a complete weighted directed graph $\graph(\nodes,\edges)$ (Eq.~\ref{eq:transfergraph}), where $\node \in \nodes$ corresponds to a single object $ \object $. Each directed edge, $\edge=(\object_i,\object_j) \in \edges$ is an ordered pair of objects $ \object_i $ and $\object_j $, where the order determines the assignment of an object to an arm $ \arm^1 $ or $ \arm^2 $. The cost of an edge $ \cost(\edge) $ is the coordinated motion planning cost of performing the transfer corresponding to $ \coma = (\object_i,\object_j) $. For instance, as shown in Fig \ref{fig:edge_matching}(middle), $\edge(A,B)$ corresponds to $ \arm^1 $ transferring `A', while $ \arm^2 $ transfers `B', and the $\cost( \edge(A,B) ) = \cost(T(\coma=(A,B))$ is the cost of such a concurrent motion. Note that these costs refer to only the transfer costs i.e., starting from the arms picking up the objects $ A $ and $ B $ at the initial poses(as shown in Fig \ref{fig:edge_matching}(left)), and moving them to their target poses. It should also be noted that since the arms are different, in general, $\cost( \edge(A,B) )$ is not necessarily equal to $\cost( \edge(B,A) )$.

Following the observation made in Eq.~\ref{eq:cost_function}, the cost of the transfers can be reasoned about independent of their order. 
Define $\scomaset$ as the unordered set of $\coma\in\scoma$, then unordered transfer cost component corresponds to $\sum_{\coma\in\scomaset}\cost(T(\coma))$.
% We defined $ \scoma $ to be an ordered sequence of object-to-arm assignments. Let $ \scoma $ be a candidate solution to the synchronized dual-arm rearrangement problem. $ \scoma $ is an ordered sequence of $ \coma $, and $ \scomaset $ denote the unordered set of all $ \coma $ that comprise $ \scoma $. 
Since $ \graph $ is a complete graph where every edge corresponds to every possible valid $ \coma $, $ \scomaset $ must also be a part of $ \graph $. 
From the construction of the graph $ \graph $, $ \scomaset \subset \edges$. By definition, a candidate solution of a monotone dual-arm rearrangement problem must transfer every object exactly once. In terms of the graph, this means that in the subset of edges $ \scomaset $ every vertex appears in only one edge. We arrive at the following crucial observation.

{\lemma [Perfect Matching]
Every candidate solution to a monotone dual-arm rearrangement problem comprises of a set of unordered object-to-arm assignments $ \scomaset $ that is also a perfect matching solution on $ \graph $.
}

As per the decomposition of the costs in Eq. \ref{eq:cost_function}, it follows that the $ \cost(\scomaset) $ corresponds to the cost of the transfers in the solution. Given the reduction to edge-matching, the solution to the minimum-weight perfect edge matching problem on such a graph would correspond to a $ \scomaset $ that optimizes the cost of \textit{transfers} of all the objects.

{\lemma [Optimal Matching]
The set of object-to-arm assignments $ \scomaset^* $ that minimizes the cost of object transfers is a solution to the minimum-weight perfect matching problem on $ \graph $.
}
% $$
% \scomaset^* = \underset{ \scomaset \in \mathtt{EM(\graph)}  }{ \mathtt{argmin} } \ \cost(\scomaset)
% $$

Once such an optimal assignment $ \scomaset^* $ is obtained, the missing part of the complete solution is the set of transits between the object transfers and their ordering. Construct another directed graph $ \tspgraph $ (Eqs. \ref{eq:transitgraph}), where the vertices comprise of the $ \coma \in \scomaset^*$. An edge $ e(\coma_{u\rightarrow v}) $ between any two vertices corresponds to the coordinated \textit{transit} motion between them. For instance, as in Fig \ref{fig:edge_matching}(right) for an edge between $ \coma(A,B) $ and $ \coma(C,D) $, $ \arm^1 $ moves from the target pose of object `A' to the starting pose of object `C', and similarly $ \arm^2 $ moves from the target of `B' to the start of `D'. An additional vertex corresponding to the starting(and ending) configuration of the two arms ($ \startq $) is added to the graph. The graph is fully connected again to represent all possible transits or moves.

A complete candidate solution to the problem now requires the sequence of~$ \coma $. By definition of the edges, the complete solution begins at~$ \startq $ and also ends there. This is a complete tour over~$ \tspgraph $, that visits all the vertices i.e., an ordered sequence of vertices $\tour = ( {\startq}, \scoma_\tour, {\startq}  )$.

{
\lemma [Tour]
Any complete tour $ \tour $ over the graph $ \tspgraph $, corresponds to a sequence of object-to-arm assignments $ \scoma_\tour $ and is a candidate solution to the synchronous dual-arm rearrangement problem.
}

% In the problem setup it was assumed that $ \trajset(\scoma) $ contained the start and end transits to $ \startq $. Additionally, the entire trajectory contained both transfers and transits. Since we have considered all the transfers in $ \cost(\scomaset^*) $, what remains is the transits. Then 
% The cost of the tour $ \tour $ can be written as $\cost(\tour) = \sum_{e(u,v)\in\tour}\cost({M}( \coma_{u\rightarrow v} ))$.

Let $ \mathbb{P}^{\scomaset} $ represent the set of all possible ordering of the elements in $ \scomaset $. This means, any candidate tour corresponds to a $ \scoma_\tour\in\mathbb{P}^{\scomaset} $. An optimal tour on $ \tspgraph $ minimizes the transit costs over the all possible candidate solutions in  $ \mathbb{P}^{\scomaset^*} $.
\vspace{-0.1in}
\begin{equation}
\scoma^+ = \underset{ \scoma \in  \mathbb{P}^{\scomaset^*}   }{ \mathtt{argmin} } \ \sum_{e(u,v)\in\tour}\cost({M}( \coma_{u\rightarrow v} ))  )
\vspace{-0.05in}
\end{equation}

This differs from the true optimal $ \scoma^* $, since the second step of finding the optimal transit tour only operates over all possible solutions that include the optimally matched transfers obtained in the first step.
The insight here is that, even though $ \scoma^+ $ reports a solution to a hierarchical optimization objective, the algorithm operates over a much smaller search space, and operates over sub-problems which are more efficient to solve than the baselines.\\

\noindent\textbf{Algorithm}: This section describes the algorithm {\tt {Tour\_Over\_Matching}} (\algo) outlined in the previous section. The steps correspond to Algo \ref{algo:tom}. 
% The algorithm takes as input the set of objects $ \objectset $ and the start(and end) configuration $ \startq $ of the arms.

\noindent $ \mathtt{transfer\_graph} $: This function constructs a directed graph $ \graph $ defined by Eqs.~\ref{eq:transfergraph}. This step creates a graph with $ n $ vertices and $ \permu[n]{2} $ edges.

\begin{wrapfigure}{r}{0.6\textwidth}\vspace{-0.3in}
  \begin{minipage}{0.6\textwidth}
  \vspace{0pt} 
    \begin{algorithm}[H]
    \caption{{\tt \algo}$ (\objectset, \startq, \ainit, \atarget) $}
    \label{algo:tom}
    $ \graph \leftarrow \mathtt{transfer\_graph}(\objectset, \ainit, \atarget) $\;
    $ \scomaset^* \leftarrow \mathtt{optimal\_matching}(\graph) $\;
    $ \tspgraph \leftarrow \mathtt{transit\_graph}(\scomaset^*, \startq) $\;
    $ \scoma^+\leftarrow\mathtt{optimal\_tour}(\tspgraph) $\;
    $ \mathbf{\mathtt{return}}\ \trajset(\scoma^+)$\;
    \end{algorithm}
  \end{minipage}
  \vspace{-0.3in}
\end{wrapfigure}

\noindent $ \mathtt{optimal\_matching} $: This function takes the graph $ \graph $ constructed as an argument and returns the unordered set of edges, corresponding to the set of optimal transfers over $ \graph $. 
\commentdel{The first step is to convert the directed graph $ \graph $ into an equivalent undirected graph $ \graph_U $. Since $ \graph $ is complete, every pair of vertices has two directed edges between them. $\graph_U$ only preserves the minimally weighted connection for every vertex pair. Optimal matching is solved using an Edmond's Blossom Algorithm~\cite{edmonds1965maximum,galil1986ev} implementation~\cite{dezsHo2011lemon}. 
% Since the algorithm performs maximum matching by default, our desired minimally weighted matching can be obtained by subtracting the edge weights from some arbitrarily large number.
}
\commentadd{Optimal matching over an undirected graph can be solved using Edmond's Blossom Algorithm~\cite{edmonds1965maximum,galil1986ev,dezsHo2011lemon}. The directed graph $ \graph $ is converted into an equivalent undirected graph $ \graph_U $. Since $ \graph $ is complete, every pair of vertices shares two directed edges. $\graph_U$ only preserves the minimally weighted connection for every vertex pair. The result of matching}
% The result 
is a subset of edges on $ \graph_U $ which correspond to a set of directed edges on $ \graph $ i.e., $ \scomaset^* $. The runtime complexity of the step is $ O(|\edges||\nodes|\log|\nodes|) = O(\permu[n]{2}n\log n ) = O(n^3\log n)$.

\noindent $ \mathtt{transit\_graph} $: This function constructs the directed transit graph $ \tspgraph $ as per the set of Eqs.~\ref{eq:transitgraph}. This constructs $ \frac{n}{2} + 1 $ vertices and $ \permu[(\frac{n}{2} + 1)]{2} $ edges.

\noindent $\mathtt{optimal\_tour}  $: Standard TSP solvers like Gurobi~\cite{gurobi} can be used to find the  
% It should be noted that a lot of efficient solvers exist for such problems, that provide various degrees of approximation. 
optimal tour over $ \tspgraph $ corresponding to $ \scoma^+ $.

The costs are deduced from coordinated motion plans over edges. The total number of such calls compared to the count from the baseline in Equ. \ref{eq:mpcount}, shows a saving in the order of $ O(n^2)$ queries ($\textrm{\# of Transfers} + \textrm{\# of Moves}$). 
% \kiril{Consider completing the calculation.}
\vspace{-0.1in}
$$
\frac{\textrm{Baseline}\ \#}{{\algo}\ \#} = 
\frac{  {\permu[n]{2}} +   \Big{(}{{\permu[n]{2}}\times\permu[n-2]{2}}\Big{)} + \Big{(}2\times{\permu[n]{2}}\Big{)}   }
{\permu[n]{2} + \permu[(\frac{n}{2} + 1)]{2}} =
\frac{4(n-1)((n-5)n+9)}{5n-2}
$$\vspace{-0.1in}

The evaluation performed here focuses on the maximum of distances (Eq.~\ref{eq:cost_function}) for a fair comparison with the other methods. The prioritization of optimization objective, however, is also amenable to other cost functions, where carrying objects is often more expensive than object-free motions. 

\section{Integration with Motion Planning}
\label{sec:integration}
The algorithms described so far are agnostic to the underlying motion planner.
Depending upon the model of the application domain, different motion planning primitives might be appropriate. For planar environments with disk robot pickers~(similar to delta robots), recent work~\cite{kirkpatrick2016characterizing} characterizes the optimal two-disk coordinated motions. The current implementation uses a general multi-robot motion planning framework \drrtstar~\cite{Dobson:2017aa} for dual-arm coordinated planning.

% It should be noted that sometimes such motion plans might not be found either because no such solution exists, or because of a limited time budget. In such cases the edge costs should either be set of some arbitrarily large value, or removed altogether. 

In practice the cost of generating and evaluating two-arm motions can dominate the overall running time of the algorithm, when compared to the combinatorial ingredients that discover the high-level plan, i.e., execution order and and arm assignment. Even though \algo reduces this, further improvements can be made with lazy evaluation. 

\begin{wrapfigure}{r}{0.6\textwidth}
  \vspace{-0.1in}
  \begin{minipage}{0.6\textwidth}
  \vspace{0pt} 
  \begin{algorithm}[H]
  \caption{{\tt Lazy\_Evaluation}$ (\mathtt{ALGO}, \mathcal{H}, \mathtt{MP}) $}
  \label{algo:lazy}
  $ \edges_{\mathtt{b}}\leftarrow\emptyset $;  $ \D \leftarrow \emptyset $\;
  \While{$ \D=\emptyset \land \mathtt{time\_not\_exceeded} $}
  {
  %	$ \mathtt{planner}\leftarrow\mathcal{H} $\;
      $ \scoma\leftarrow \mathtt{ALGO} ( \mathcal{H},   \edges_{\mathtt{b}}  ) $\;
  %	$ \mathtt{planner}\leftarrow\mathtt{MP} $;
%       $ \D \leftarrow \emptyset $\;
      \For{$ \coma_i, \coma_{i\rightarrow{i+1}} \in \scoma$}
      {
          $ \D_i,\D_{i\rightarrow i+1} \leftarrow \mathtt{MP}(\coma_i), \mathtt{MP}( \coma_{i\rightarrow{i+1}} ) $\;
          $ \D\leftarrow (\D,\D_i,\D_{i\rightarrow i+1} ) $\;
          \If{$ \D_i = \emptyset $}
          {
              $ \edges_{\mathtt{b}} \leftarrow \edges_{\mathtt{b}}\cup \{\coma_i\}$; $ \D \leftarrow \emptyset$\;
          }
          \If{$ \D_{i\rightarrow i+1} = \emptyset $}
          {
              $ \edges_{\mathtt{b}} \leftarrow \edges_{\mathtt{b}}\cup \{\coma_{i\rightarrow{i+1}}\}$; $\D \leftarrow \emptyset $\;
          }
          \lIf{$ \D=\emptyset $}
          {
          $\mathbf{break}$
          }
      }
  }
  $\mathbf{return}\ D$
  \end{algorithm}
  \end{minipage}
  \vspace{-0.35in}
\end{wrapfigure}

% \kiril{Are we referring to the correct paper here? }. 
% This heuristic corresponds to the pre-processed shortest path costs over underlying roadmaps. 
% \kiril{I suggest to replace this sentence with a high-level description omitting the term "underlying roadmaps": "This heuristic corresponds to pre-processed lower estimates of the shortest-path costs for the two arms."}
% The online part of the estimate is tantamount to a simple look-up. 

\noindent\textbf{Lazy Evaluation}: Recent work~\cite{shome2017improving} introduces heuristics for $ \drrtstar $, which pre-process estimates of the shortest path costs for the arms. Dual-arm rearrangement can be significantly sped up if the motion planning queries are replaced with look-ups of such heuristics. Once a candidate $ \scoma $ is obtained, motion planning can evaluate the solution $ \trajset(\scoma) $. If this fails, the algorithm tries other sequences. 
% \kiril{In line 8 of the algorithm, the first "$D$" in the curly brackets is redundant, right?}

The algorithm Algo \ref{algo:lazy} takes as input the algorithm $ \mathtt{ALGO} $, a heuristic $ \mathcal{H} $, and a motion planner $ \mathtt{MP} $. $ \edges_{\mathtt{b}}$ keeps track of the blocked edges. The process keeps generating candidate solutions using the $ \mathtt{ALGO} $ (Line 3). Line 5 motion plans over the candidate solution, and appends to the result (Line 6). Any failures are recorded in  $ \edges_{\mathtt{b}}$ (Lines 8,10), and inform subsequent runs of $ \mathtt{ALGO} $.

\noindent\textbf{Completeness}: 
% The heuristic versions would be an approximation of the original optimization guarantees of the algorithms. It should be noted that this framework can be used with all the dual-arm approaches ie., exhaustive search, MILP and \algo. 
The lazy variants give up on optimality for the sake of efficiency but given enough retries the algorithms will eventually solve every problem that {\tt ALGO} can. Since the motion planning happens in the order of execution, the object non-interactivity assumption is relaxed.

\noindent\textbf{Smoothing}: 
% The solutions obtained out of the methods adhere to the defined synchronization assumptions of the problem. 
Applying velocity tuning over the solution trajectories for the individual arms relaxes the synchronization assumption
by minimizing any waits that might be a by-product of the synchronization. 
Smoothing does not change the maximum of distances, only improves execution time. Indications that the smoothed variants of the synchronous solutions do not provide significant savings in execution time are included in the Appendix for the interested reader.

It should be noted that in an iterative version of \algo, in order to explore variations in $ \scomaset^* $ if failures occur in finding $\tour$, some edges need to be temporarily blocked on $ \scomaset^* $. The search structures can also be augmented with {\tt NO\_ACT} tasks, for possible $\coma$ where one of the arms do not move. 
% This effectively includes single arm solutions in the search as well.

\section{Bounding Costs under Planar Disc Manipulator Model}
\label{sec:cost_bounds}

Following from Lemma \ref{lem:transferdomination}, the current analysis studies the dual arm costs in the randomized unit tabletop setting, with $c_t$ as the cost measure per unit distance. 
For the $2$-arm setting, assume for simplicity that each arm's volume is represented as a disc of some radius 
$r$. For obtaining a $2$-arm solution, we partition the $n$ objects randomly 
into two piles of $\frac{n}{2}$ objects each; then obtain two initial solutions 
similar to the single arm case. It is expected (Eq.~\ref{eq:single-cost-simple}) that these two halves 
should add up to approximately $(c_{pd} + 0.52c_t)n$.

From the initial $2$-arm solution, we construct an {\em asynchronous} $2$-arm 
solution that is collision-free. Assume that pickups 
and drop-offs can be achieved without collisions between the two arms, which 
can be achieved with properly designed end-effectors. The main overhead 
is then the potential collision between the two (disc) arms during transfer 
and move operations. Because there are $\frac{n}{2}$ objects for each arm to work 
with, an arm may travel a path formed by $n + 1$ straight line 
segments. Therefore, there are up to $(n + 1)^2$ intersections between the 
two end-effector trajectories where potential collision may happen. However,
because for the transfers and transits associated with one pair of objects (one 
for each arm) can have at most four intersections, there are at most 
$2n$ potential collisions to handle. For each intersection, let one 
arm wait while letting the other circling around it, which incurs a cost that is bounded by $2\pi \cdot
r \cdot c_t$.

Adding up all the potential extra cost,  
% that a $2$-arm solution has 
a cumulative cost is obtained as 
% \begin{align}\label{eq:dual-cost-distance} 
% C_{dual} = C_{single} + 2n(2\pi rc_t) \approx (c_{pd} + 0.52c_t + 4\pi rc_t)n.
% \end{align}

{\centerline
{
$C_{\rm dual} = C_{\rm single} + 2n(2\pi rc_t) \approx (c_{pd} + 0.52c_t + 4\pi rc_t)n\;.$
}
}

\noindent For small $r$, $C_{\rm dual}$ is almost the same as $C_{\rm single}$ 
$c_t$ is a distance (e.g., energy) cost. Upon considering the maximum of the two arc lengths or makespan (Eq.~\ref{eq:cost_function}),
the $2$-arm cost becomes $C_{\rm dual}^t \approx (c_{pd} + 0.52c_t)\frac{n}{2} + 4n\pi rc_t$.
% \vspace{-0.1in}
% \begin{equation}\label{eq:dual-cost-time}
% C_{dual}^t \approx (c_{pd} + 0.52c_t)\frac{n}{2} + 4n\pi rc_t
% \vspace{-0.1in}
% \end{equation}
% {\centerline
% {
% $C_{dual}^t \approx (c_{pd} + 0.52c_t)\frac{n}{2} + 4n\pi rc_t$
% }
% }
The cost ratio is
\vspace{-0.1in}
\begin{equation}\label{eq:makespan-ratio}
\frac{C_{\rm dual}^t}{C_{\rm single}} \approx 
\frac{(c_{pd} + 0.52c_t)\frac{n}{2} + 4n\pi rc_t}{(c_{pd} + 0.52c_t)n}
= \frac{1}{2} + \frac{4\pi rc_t}{c_{pd} + 0.52c_t}\;.
% \vspace{-0.1in}
\end{equation}

\commentdel{
 \begin{wrapfigure}{r}{1.4in}
  \vspace{-.45in}
	\centering
    \includegraphics[width=1.4in]{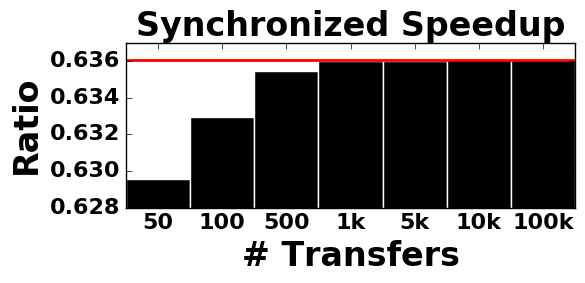}
    \vspace{-.4in}
	\caption{Empirical cost ratio versus the estimate}
  	\label{fig:bounds}
    \vspace{-.4in}
\end{wrapfigure}
}

When $r$ is small or when $\frac{c_t}{c_{pd}}$ is small, the $2$-arm 
solution is roughly half as costly as the single arm solution. On 
the other hand, in this model a $2$-arm solution does not do better than $\frac{1}{2}$ of 
the single arm 
solution. This argument can be extended to $k$-arms as well.
\vspace{-0.08in}
\commentadd{
\begin{theorem}
For rearranging objects with non-overlapping starts and goals that are
uniformly distributed in a unit square,  a $2$-arm solution can have an 
asymptotic improvement of $\frac{1}{2}$ over the single arm solution. 
\label{thm:karm}
\end{theorem}
}
The synchronization assumption changes the expected cost of the solution. The random partitioning of the $n$ objects into two sets of $\frac{n}{2}$ object with a random ordering of the objects yields $\frac{n}{2}$ pairs of objects transfers, which dominate the total cost for large $n$. The cost~(Eq.~\ref{eq:cost_function}) of $\frac{n}{2}$ synchronized transfers ($\coma_i$) includes $\frac{n}{2}c_{pd}$ and $C^{\rm sync}_{sg} \approx (\mathtt{E}(max(l_1,l_2))c_t)\frac{n}{2}$, where $\mathtt{E}(max(l_1,l_2))$ is the expected measure of the max of lengths $l_1$,$l_2$ of two randomly paired transfers. Using the \textit{.jpg}\cite{ghosh1951random} of lengths of random lines in an unit square and integrating over the setup, results in the value of $\mathtt{E}(max(l_1,l_2))$ to be $0.66$. 
% (Section \ref{sec:appendix}). 
This means $C_{\rm dual}^{\rm sync} \approx (c_{pd} + 0.66c_t)\frac{n}{2} + 4n\pi rc_t$.
% \vspace{-0.1in}
% \begin{equation}\label{eq:synchronized-cost}
% C_{dual}^{sync} \approx (c_{pd} + 0.66c_t)\frac{n}{2} + 4n\pi rc_t
% \vspace{-0.1in}
% \end{equation}
% {\centerline
% {
% $C_{dual}^{sync} \approx (c_{pd} + 0.66c_t)\frac{n}{2} + 4n\pi rc_t$
% }
% }
The synchronized cost ratio is
\vspace{-0.1in}
\begin{equation}\label{eq:synchronized-ratio}
\frac{C_{\rm dual}^{\rm sync}}{C_{\rm single}} \approx 
\frac{(c_{pd} + 0.66c_t)\frac{n}{2} + 4n\pi rc_t}{(c_{pd} + 0.52c_t)n}
= \frac{1}{2} + \frac{ (0.07 + 4\pi r)c_t}{c_{pd} + 0.52c_t}\;.
\vspace{-0.1in}
\end{equation}

When $\frac{c_t}{c_{pd}}$ is small, even the synchronized $2$-arm solution provides an improvement of $\frac{1}{2}$. For the case when both $r$ and $c_{pd}$ are small, we observe that the ratio approaches $0.636$. 
\commentdel{Fig \ref{fig:bounds} confirms empirically that in randomized trials on an unit square, the ratio of $\frac{C_{dual}^{sync}}{C_{single}}$ when $c_{pd}=0,r=0$, converges to the calculated estimate (red line).}
\vspace{-0.08in}
\begin{theorem}
For rearranging objects with non-overlapping starts and goals that are 
uniformly distributed in a unit square,  a randomized $2$-arm synchronized solution can have an 
asymptotic improvement of $\frac{1}{2}$ over the single arm solution if $\frac{c_t}{c_{pd}}$ is small, and a improvement $\approx 0.64$ when both $c_{pd}$ and $r$ are small. 
\end{theorem}

\commentadd{
\noindent \textbf{Note on bounds:} The proposed simplified model does not apply immediately to general configuration spaces, where the costs and collision volumes do not have the same nice properties of a planar disk setup. The experimental section, however, indicates that the benefits and speedups exist in these spaces as well.
}
% \rahul{this paragraph is phrased contradictorily. It makes sense if the first sentence is for asynchronous cases.}
% We note that the same can be said for a {\em synchronous} $2$-arm solution if
% when $c_t/c_{pd}$ is small. 
% Under the synchronization assumption, $C_{dual}^t$
% is no longer necessarily expected to be $\frac{C_{single}}{2}$ for large $n$. 
% Experiments in the next section indicate that improvements over single arm solutions in practical settings.

%\textcolor{red}{NOTE: We can also generate plots illustrating how the ratio 
%change as $n$, $r$, $c_{pd}:c_t$ changes. We may also further generalize the 
%$r$ based collision term to a more general collision term.}

%\begin{figure}[h]
%%  \begin{center}
%\centering
%    \includegraphics[width=2in]{figures/bounds}
%%  \end{center}
%  \caption{Bounds}
%  	\label{fig:bounds}
%\end{figure}

\section{Evaluation}
\label{sec:evaluation}
%\textit{Describe experiments: we will design benchmarks per model
%Describe alternative algorithms }
%a) \textit{single-arm solution per model,}
%b) \textit{exhaustive search integrated with the corresponding path planner for each model,}
%c) \textit{approximate solution we propose [unless we come up with something
%  better for the easier models].}
%d) \textit{1-TSP broken into 2 and coordinated}
%
%\textit{What do we measure: a) solution quality in terms of the cost metrics
%identified and b) computation time.}

\begin{wrapfigure}{r}{2.2in}
    \vspace{-.3in}
	\centering
		\includegraphics[width=1.1in,trim={1cm 9cm 1cm 5cm},clip]{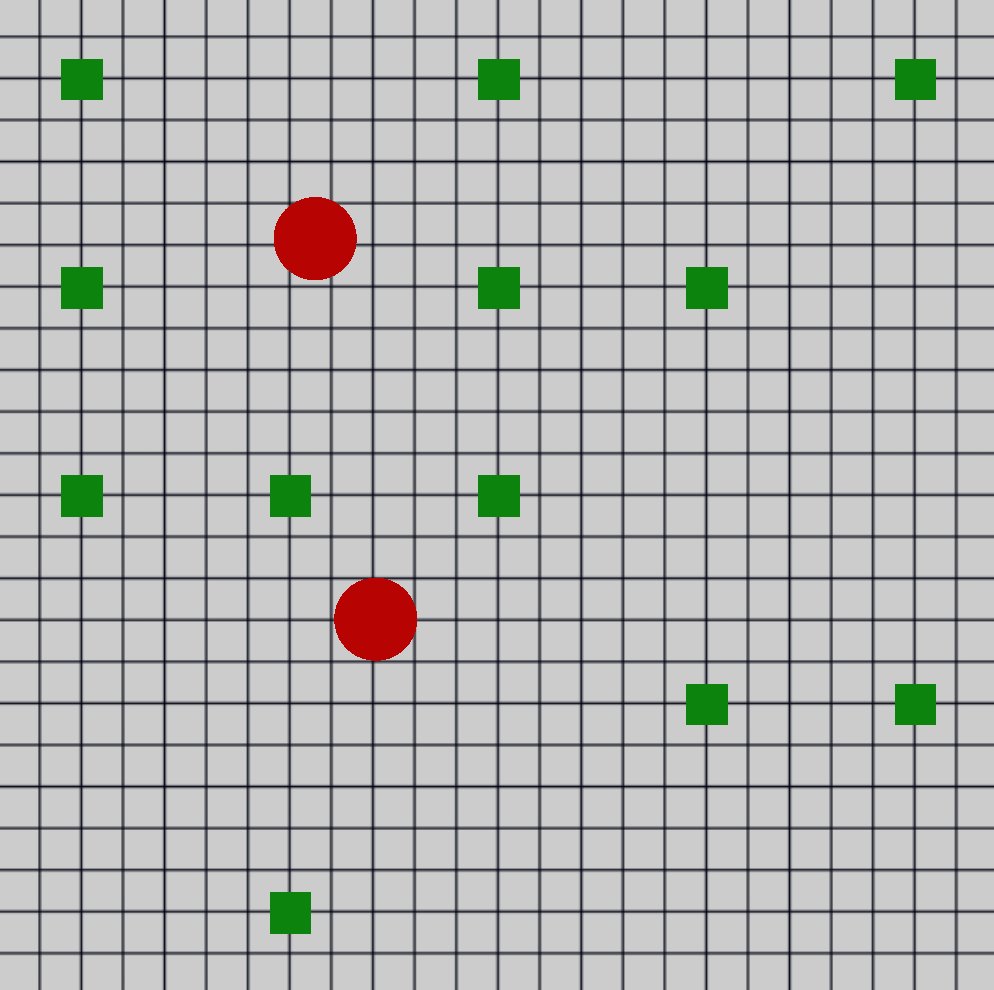}
		\includegraphics[width=1.05in]{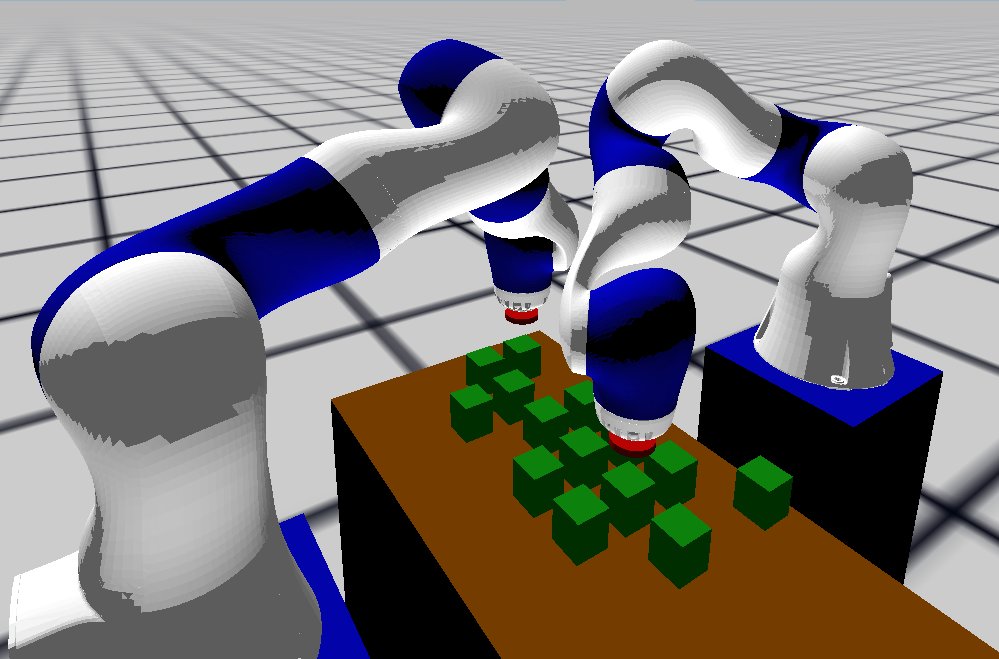}
		\vspace{-.3in}
		\caption{\textit{Picker} and \textit{Manipulator} trials.}
		\label{fig:benchmarks}
    \vspace{-.25in}
\end{wrapfigure}
This section describes the experiments performed to evaluate the algorithms in two  domains shown in Fig \ref{fig:benchmarks}: a) simple picker and b) general manipulators.
%\begin{figure}[ht]
%	\centering
%	\includegraphics[width=1.9in]{figures/simple_picker_benchmark}
%	\includegraphics[width=1.8in]{figures/kuka_benchmark2}
%	\caption{The benchmarks: \textit{Simple Picker} and the \textit{General Manipulators} }
%	\label{fig:benchmarks}
%\end{figure}
In order to ensure monotonicity, the object starts and goals do not overlap. Uniform cuboidal objects simplify the grasping problem, though this is not a limitation of the methods. $ 50 $ random experiments were limited to $300s$ of computation time. The underlying $ \drrtstar $ motion planner is restricted to a max of $ 3s $ per plan.
A comparison point includes a random split method, which splits $ \objectset $ at random into two subsets and chooses an arbitrary ordering. Maximum of distances cost is compared to the single arm solution~\cite{193}. Computation times and success rates are reported. The trends in both experiments show that in the single-shot versions, exhaustive and \milp tend to time-out for larger $n$. Lazy variants scale much better for all the algorithms, and in some cases increase the success ratio due to retries. \algo has much better running time than exhaustive and \milp, and producing better and more solutions than random split. Overall, the results show a) our \milp succeeds more within the time limit than exhaustive, b) \algo scales the best among all the methods, and c) the cost of solutions from \algo is close to the optimal baseline, which is around half of the single arm cost.
% to demonstrate the gains from using multiple arms.  Overall, MILP scales better than exhaustive search, while finding optimal solutions. \algo scales better than the other methods, and finds high quality solutions. The lazy variants improve all the dual-arm methods.

\begin{figure}[h]
\vspace{-0.2in}
	\centering
	\includegraphics[width=3.5in]{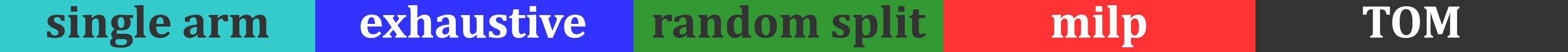}
	\includegraphics[width=2.25in]{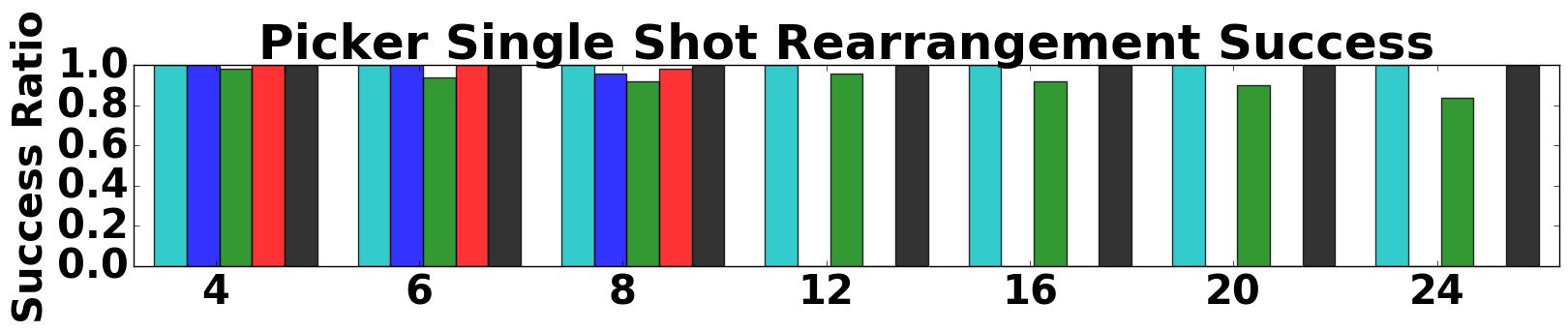}
	\includegraphics[width=2.25in]{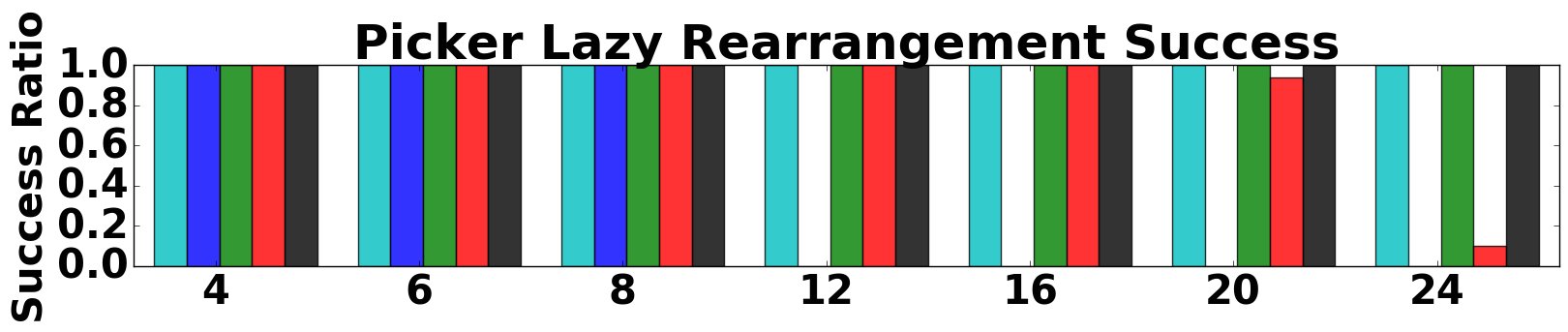}
	\includegraphics[width=2.25in]{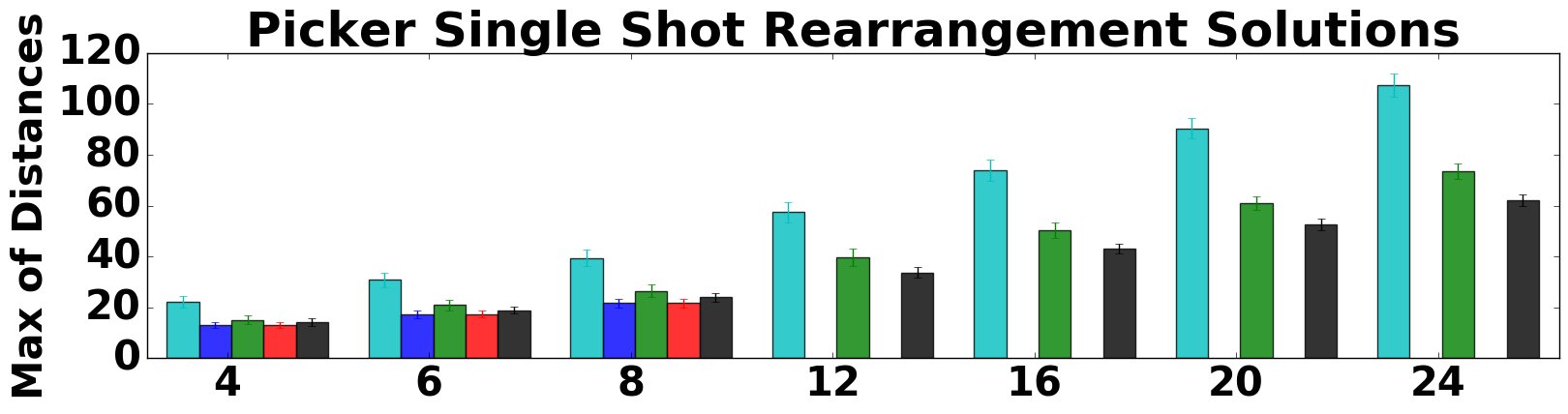}
	\includegraphics[width=2.25in]{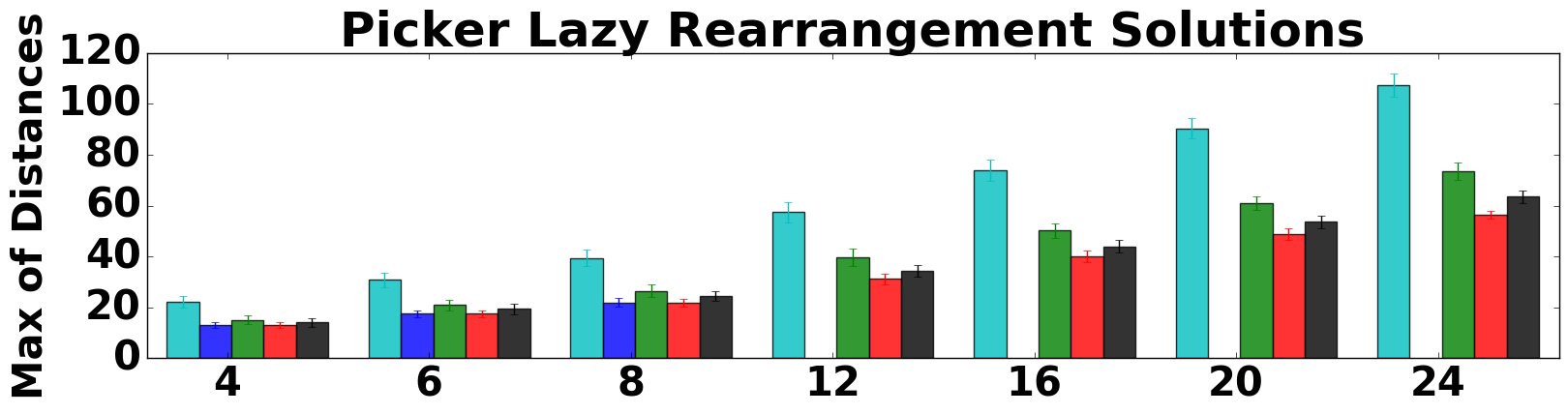}
	\includegraphics[width=2.25in]{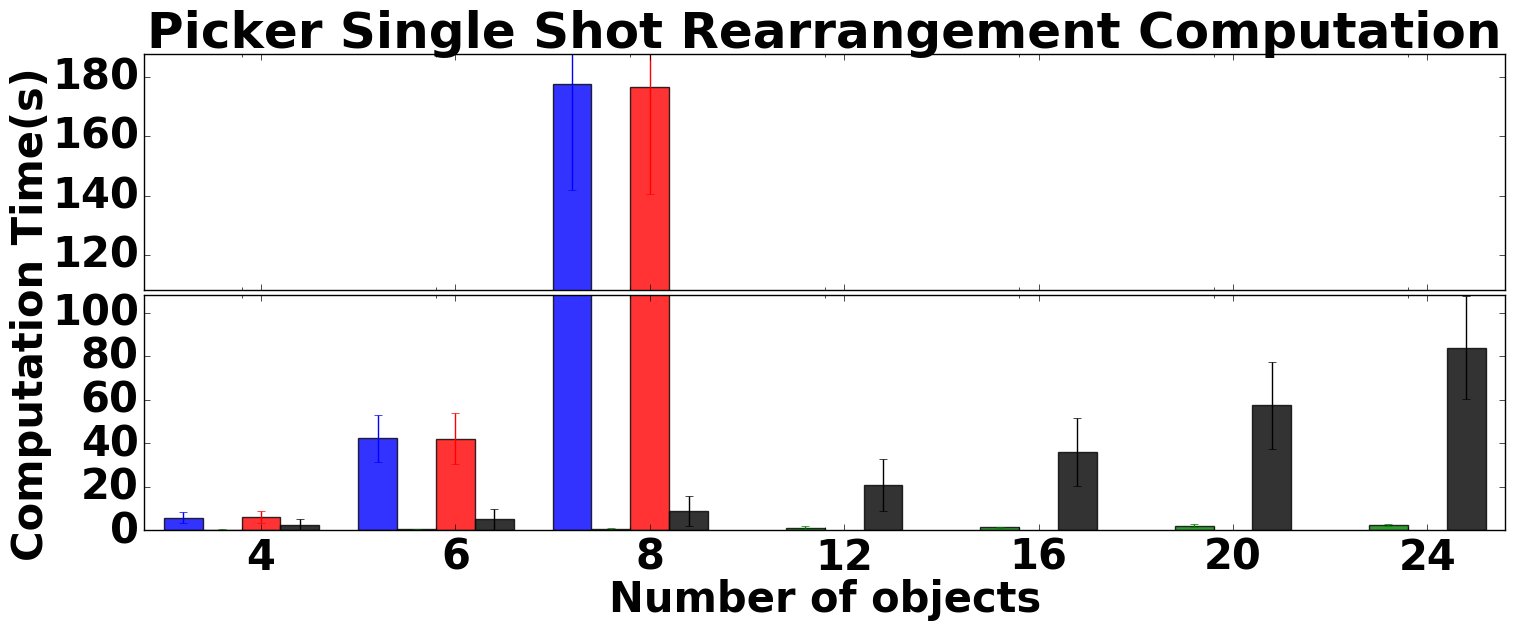}
	\includegraphics[width=2.25in]{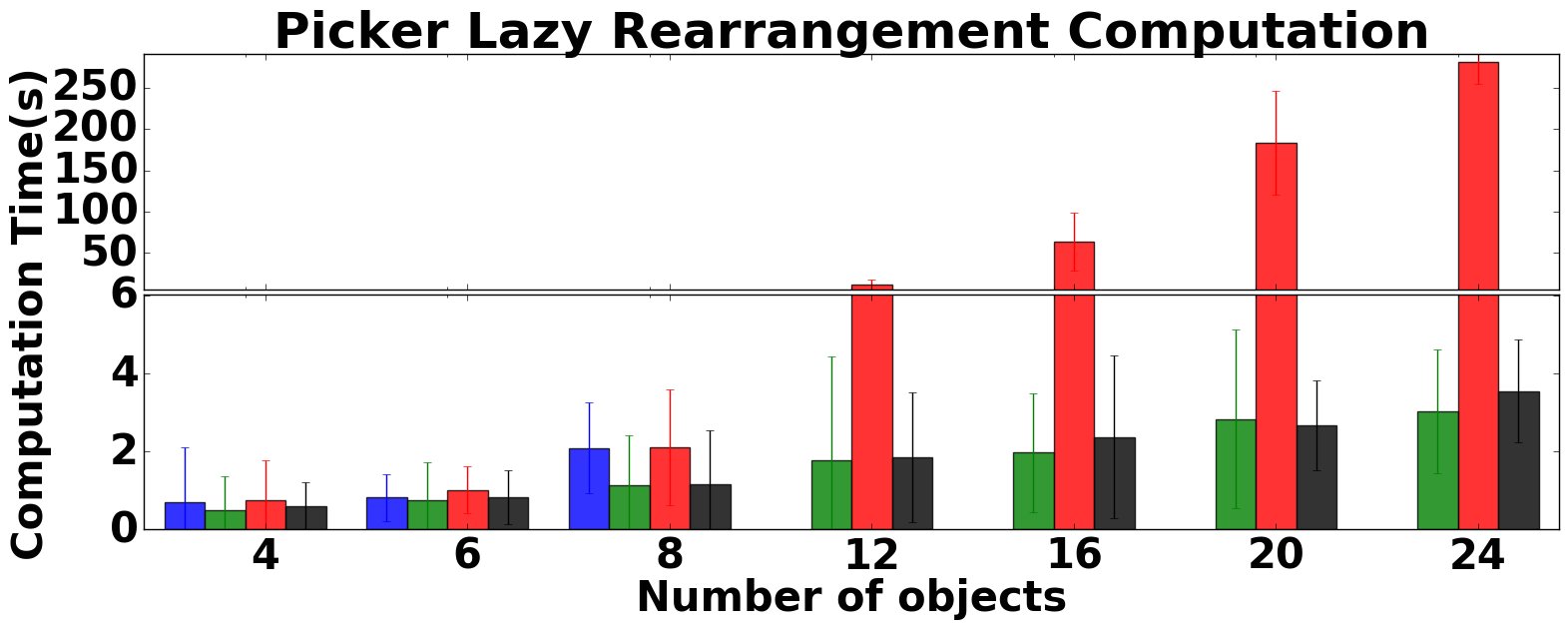}
	\vspace{-0.15in}
	\caption{\textit{Simple Picker} results with success\textit{(top)}, solution costs\textit{(middle)}, and computation\textit{(bottom)} reported for single-shot\textit{(left)} and lazy\textit{(right)} versions of the methods}
    \vspace{-0.3in}
	\label{fig:disk}
\end{figure}

%\subsection{Simple Picker}
%The results are shown in Figure~\ref{fig:disk}.
\textbf{Simple Picker}: This benchmark evaluates two disk robots hovering over a planar surface scattered with objects. The robots are only free to move around in a plane parallel to the resting plane of the objects, and the robots can pick up objects when they are directly above them. Fig~\ref{fig:disk}(\textit{top}) all runs up to $ 24 $ objects succeeded for \algo. \milp scales better than exhaustive. Lazy random split succeeds in all cases~(\textit{bottom}). In terms of solution costs~(\textit{middle}) exhaustive finds the true optimal. \milp matches exhaustive and \algo is competitive. In all experiments, \algo enjoys a success rate of 100\%.
% \kiril{What is the reason for low success rate of exhaustive and milp? Is it due to exceeding the time budget?} \kiril{This paragraph misses a statement that demonstrates the benefits of our approach: "In all experiments, \algo enjoys a success rate of \%100, while having much better running time than exhaustive and milp, and producing a higher cost solution than random split. Furthermore, \algo produces a plan whose cost is around half of the single arm plan cost." Similar statements should be added to the other experiments.}

%\subsection{General Manipulator}
%The results are shown in Figure~\ref{fig:kuka}.

\begin{figure}[h]
\vspace{-0.2in}
	\centering
	\includegraphics[width=3.5in]{figures/results/labels.jpg}
	\includegraphics[width=2.25in]{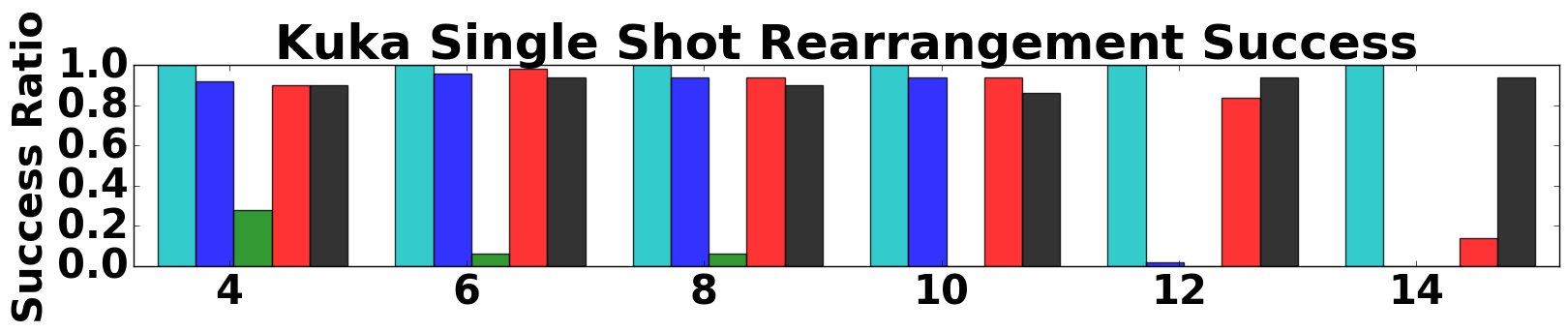}
	\includegraphics[width=2.25in]{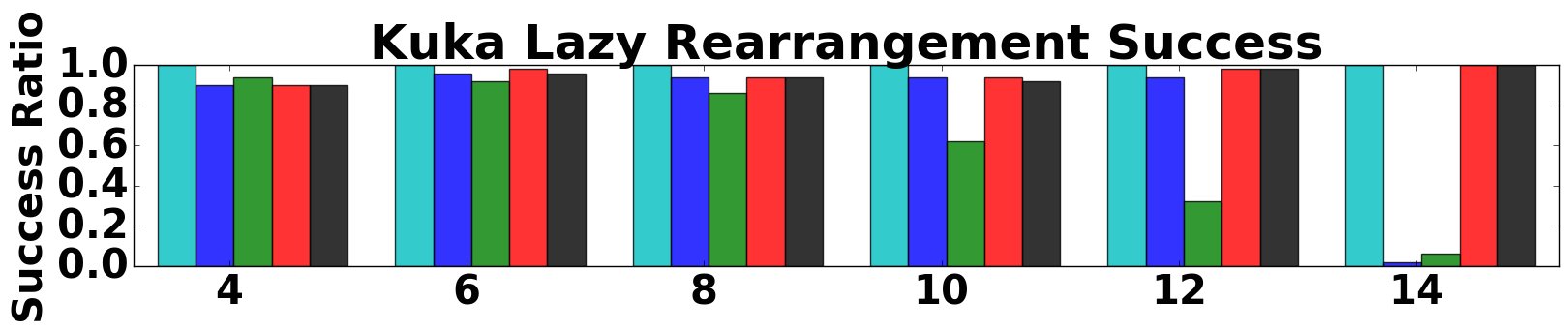}
	\includegraphics[width=2.25in]{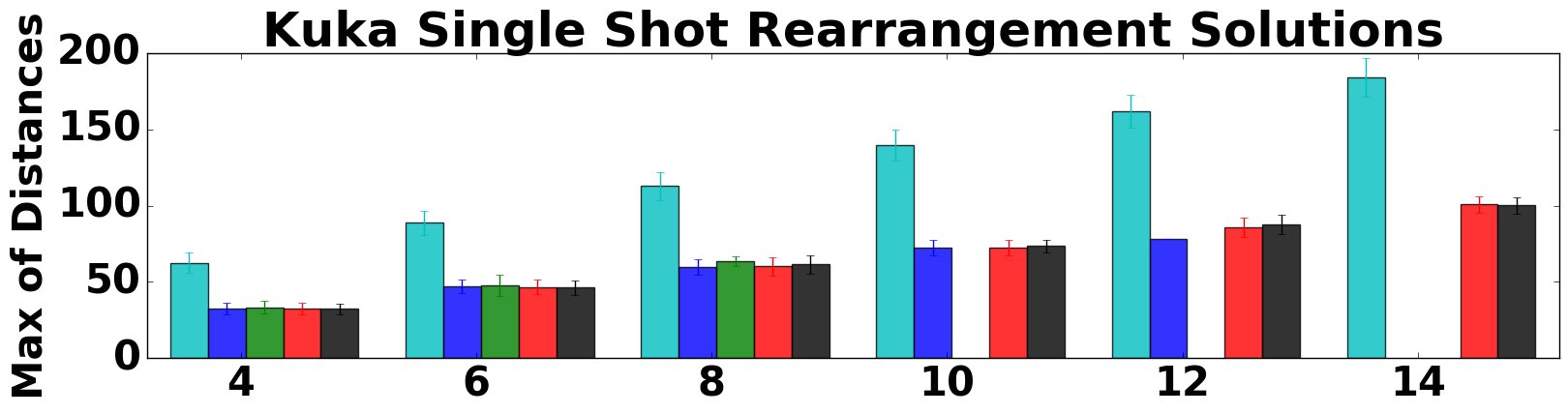}
	\includegraphics[width=2.25in]{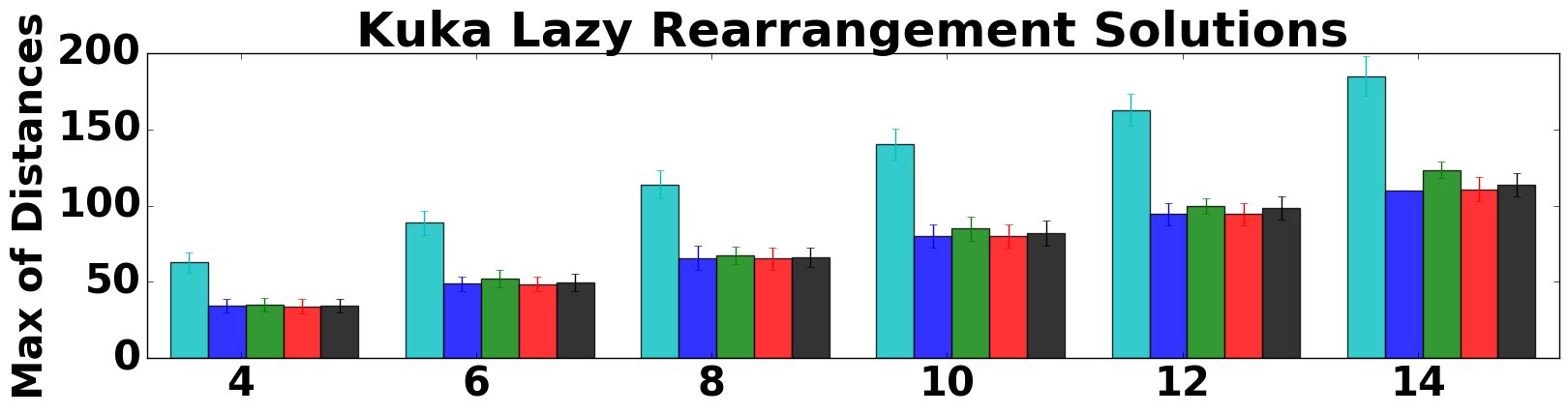}
	\includegraphics[width=2.25in]{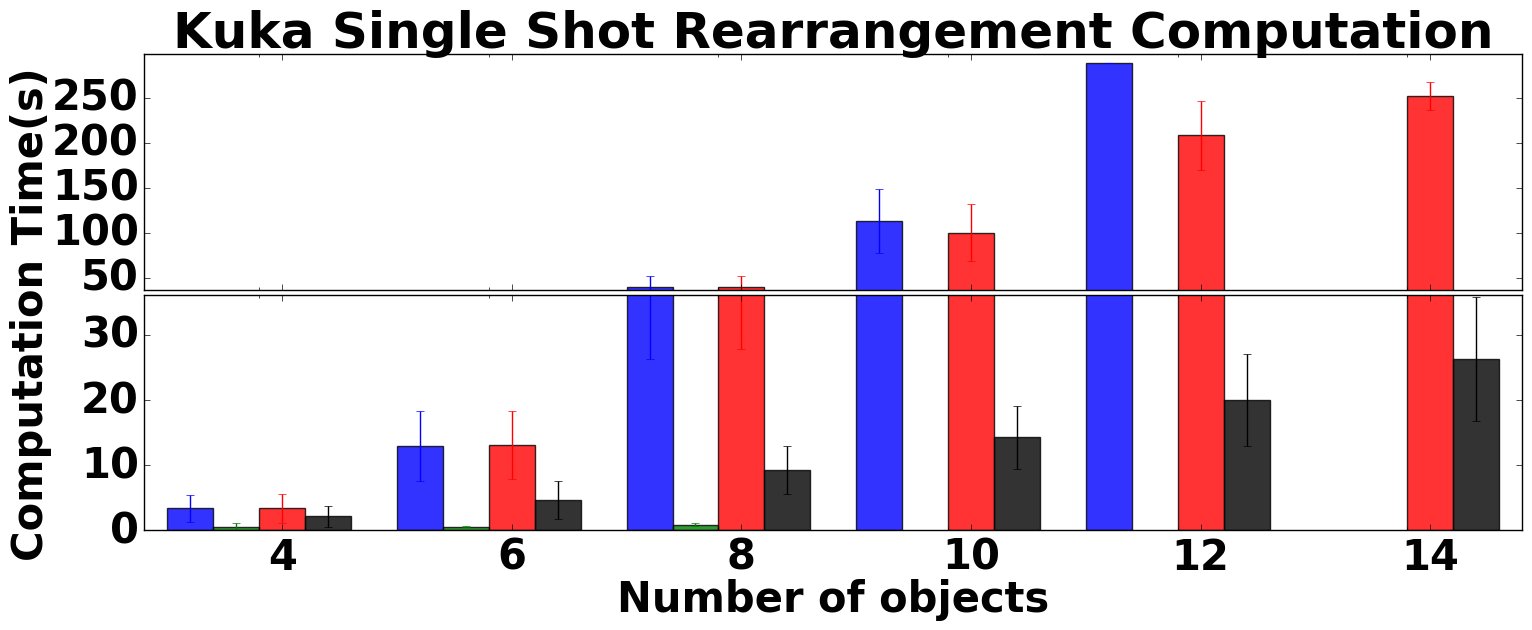}
	\includegraphics[width=2.25in]{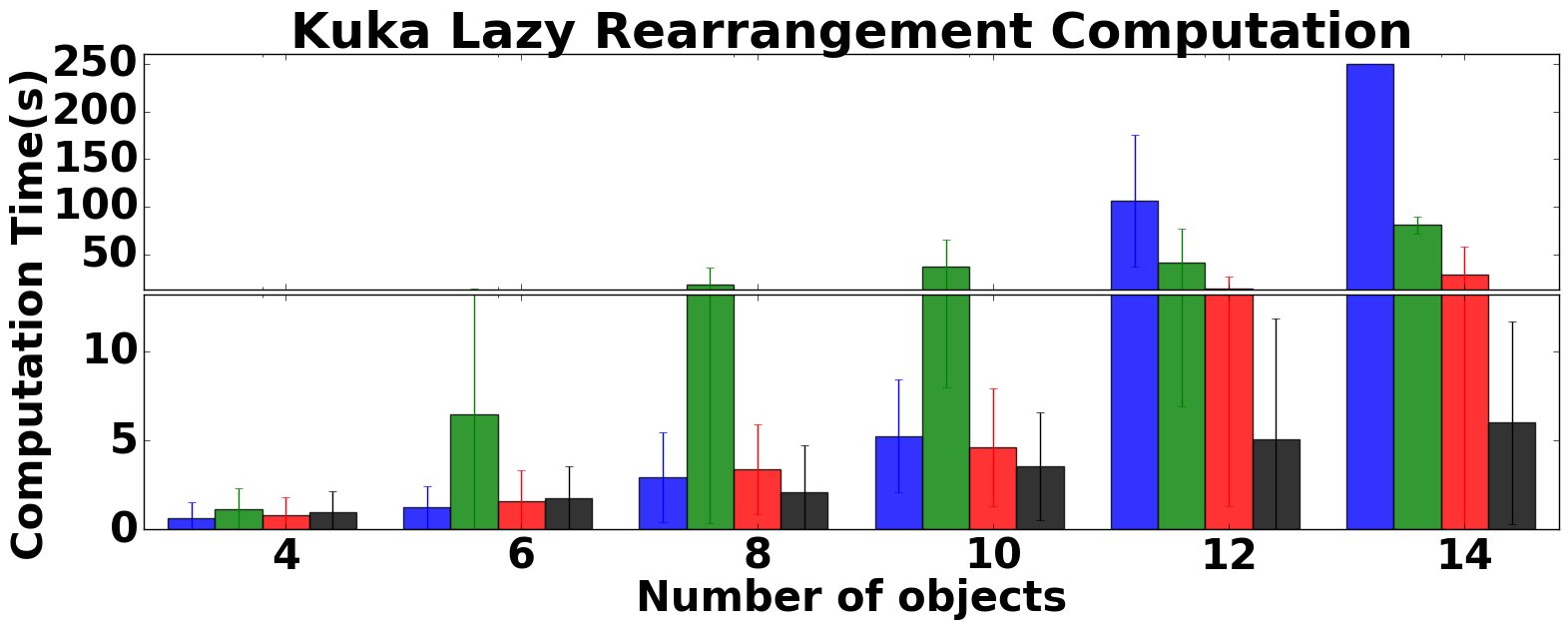}
	\vspace{-0.15in}
    \caption{\textit{\kuka}results with success\textit{(top)}, solution costs\textit{(middle)}, and computation\textit{(bottom)} reported for single-shot\textit{(left)} and lazy\textit{(right)} versions of the methods}
    \vspace{-0.3in}
	\label{fig:kuka}
\end{figure}

\textbf{General Dexterous Manipulator}: The second benchmark sets up two \kuka arms across a table with objects on it. The objects are placed in the common reachable part of both arms' workspace, and only one top-down grasping configuration is allowed for each object pose. Here (Fig~\ref{fig:kuka}) a larger number of motion plans tend to fail, so the single shot variants show artifacts of the randomness of \drrtstar in their success rates. 
% The lazy iterative variants show the robust scalability. 
Random split performs the worst since it is unlikely to chance upon valid motion plans. Single shot exhaustive and \milp scale poorly because of expensive motion planning.
Interestingly, motion planning infeasibility reduces the size of the exhaustive search tree.
The solution costs \textit{(middle)} substantiate benefits of the use of two arms. The computation times \textit{(bottom)} again show the scalability of \algo, even compared to random split.
% In terms of cost. MILP scales to more objects but the computational overhead is more than \algo. Dual-arm solutions are significantly better than single arm.

\section{Discussion}
\label{sec:discussion}
The current work demonstrates the underlying structure of synchronized dual-arm rearrangement and proposes an \milp formulation, as well as a scalable algorithm \algo that provides fast, high quality solutions. Existing efficient solvers for reductions of the dual-arm problem made \algo effective. Future work will attempt to explore the $k-$arm case and instances of non-monotone rearrangement. The incorporation of manipulation and regrasp reasoning can extend these methods to more cluttered domains.

% It can be shown however that efficient approximate solvers for more than 2 arms would be hard to design.
% Let us assume a $ \rho $-Approx algorithm, $\mathbb{A}$ exists for the problem of rearrangement with 3 arms. Assume a hypothetical manipulator that incurs a constant transit cost on a bounded plane. The problem of an optimal set of object assignments reduce to 3D-Matching. $ \rho $-Approx $\mathbb{A}$ should return a solution to 3DM. Since 3DM is hard to approximate, $\mathbb{A}$ does not exist.\kiril{The connection with 3DM is not immediate, and I don't think that there's enough information provided here in order to make sense of it. I suggest to omit this description, especially if space is limited.}

%
% The file named.bst is a bibliography style file for BibTeX 0.99c
{\small
\bibliographystyle{spmpsci}
\bibliography{manip.bib}}

\newpage

\section{Appendix}
\label{sec:appendix}
% \usepackage{amsmath,amsfonts,amssymb,mathrsfs}
% \usepackage{graphicx}
% \usepackage{psfrag,graphicx,epsfig,epsf}
% \usepackage[ruled,linesnumbered, noend]{algorithm2e}
% \usepackage{fancyhdr}
% \usepackage{wrapfig}
% \usepackage{subfigure}
% \usepackage{mathtools}
% \usepackage{url}
% \usepackage{color}
% \usepackage{verbatim}
% \usepackage{xspace}

% \DeclareMathOperator*{\argmin}{arg\,min}

% \newcommand\permu[2][^n]{\prescript{#1\mkern-2.5mu}{}P_{#2}}
% \newcommand\combi[2][^n]{\prescript{#1\mkern-0.5mu}{}C_{#2}}

% \begin{document}

% \title{Fast High-Quality Dual-Arm Rearrangement in Synchronous, Monotone Tabletop Setups: Appendix}
% %\titlerunning{Fast Methods for Synchronized Tabletop Dual-Arm Rearragement}
% \author{Rahul Shome$^1$ \and Kiril Solovey$^2$ \and Jingjin Yu$^1$ \and Kostas Bekris$^1$ \and Dan Halperin$^2$}
% %\authorrunning{Shome et al.}
% \institute{$^1$Rutgers University, NJ, USA and $^2$Tel Aviv University, Israel}

% \maketitle
% \input{notation}
% \documentclass{llncs}

%####################################################################################
%####################################################################################
%####################################################################################
%####################################################################################
%####################################################################################

\subsection{Expected k-arm cost bounds in a planar disk manipulator model}
\commentadd{
The arguments made in Theorem \ref{thm:karm} can be extended to $k$ disc arms.
In the planar unit-square setting, with $k$ arms, there are $\frac{n}{k}$ objects 
for each arm to work with. Consider the transfers and transits of a set of $k$ 
objects, one for each arm. By \cite{ChiHanYu2018WAFR}, the arbitrary rearrangement of $k$ discs 
can be achieved in a bounded region with a perimeter of $O(kr)$. 
Clearly, the per robot additional (makespan or distance) cost is bounded by some 
function $f(k, r)c_t$, which goes to zero as $r$ goes to zero. Adding up all the potential 
extra cost,  a $k$-arm solution has a cumulative cost

{\centerline
{
$C_{\rm k{\text-}arm} = C_{\rm single} + nf(k,r)c_t \approx (c_{pd} + 0.52c_t + f(k,r)c_t)n\;.$
}
}

\noindent For fixed $k$ and small $r$, $C_{\rm k{\text-}arm}$ is almost the same as $C_{\rm single}$.
%, $c_t$ is a distance (e.g., energy) cost. 
Upon considering the maximum of the two arc lengths or makespan,
the $k$-arm cost becomes $C_{\rm k{\text-}arm}^t \approx (c_{pd} + 0.52c_t)\frac{n}{k} + nf(k,r)c_t$.

The cost ratio is
\vspace{-0.1in}
\begin{equation}\label{eq:kmakespan-ratio}
\frac{C_{\rm k{\text-}arm}^t}{C_{\rm single}} \approx 
\frac{(c_{pd} + 0.52c_t)\frac{n}{k} + nf(k,r)c_t}{(c_{pd} + 0.52c_t)n}
= \frac{1}{k} + \frac{f(k,r)c_t}{c_{pd} + 0.52c_t}\;.
% \vspace{-0.1in}
\end{equation}

When $r$ is small or when $\frac{c_t}{c_{pd}}$ is small, the $k$-arm 
solution is roughly $\frac{1}{k}$ as costly as the single arm solution. On 
the other hand, in this model a $k$-arm solution does not do better than $\frac{1}{k}$ of 
the single arm.

\begin{theorem}
For rearranging objects with non-overlapping starts and goals that are
uniformly distributed in a unit square,  a $k$-arm solution can have an 
asymptotic improvement of $\frac{1}{k}$ over the single arm solution. 
\end{theorem}
}
% \rahul{Need to verify this line of reasoning}

\subsection{Expected measure of the maximum of lengths of two random lines on an unit square}
Prior work \cite{ghosh1951random} defines the \textit{.jpg} of lengths($l$) of randomly sampled lines in a rectangle of sides $a,b, \ \ a\geq b$ as 
\begin{align*}
p(l) &= (\frac{4l}{a^2b^2})\phi(l)\\
\phi(l)&= \frac{1}{2}\pi a b - a l - b l + \frac{1}{2} l^2, \ \ l\in[0,b]\\
\phi(l)&= ab \sininv(\frac{b}{l}) + a. \sqrt[]{(l^2-b^2)} - al - \frac{1}{2}b^2, \ \ l\in[b,a]\\
\phi(l)&= ab\{\sininv(\frac{b}{l})-\cosinv(\frac{a}{l})\} + a\sqrt[]{(l^2-b^2)} + b\sqrt[]{(l^2-a^2)} \\
&- \frac{1}{2}(l^2+a^2+b^2),\ \ l\in[a,\sqrt[]{(a^2+b^2)}]
\end{align*}

In the unit square model, $a=b=1$. Substituting the values, the \textit{.jpg} becomes
\begin{align}
p(l)&= 2\pi l - 8\pi l^2 + 2l^3, \ \ l\in[0,1]\\
p(l)&= 4l\sininv(\frac{1}{l}) - 4l\cosinv(\frac{1}{l}) + 8l\sqrt[]{(l^2-1)} -2l^3 -4l, \ \ l\in[1,\sqrt[]{2}]
\end{align}

Assuming two random sets of lines, representing transfers in a random split of objects between two arms, we need the expected value of the maximum of these pairwise lengths ie., $\mathtt{E}(max(l_1,l_2)), \ \ l_1,l_2\ i.i.d, \ \ l_1 \sim p, l_2 \sim p$.
This is estimated using the \textit{.jpg} obtained. 
\begin{equation}
\mathtt{E}(max(l_1,l_2)) = \int_0^{\sqrt[]{2}} \int_0^{\sqrt[]{2}} max(l_1,l_2)p(l_1)p(l_2) dl_2 dl_1 \approx 0.663
\end{equation}
The result is calculated by taking into account the combination of different ranges of $p(l)$ and $max(l_1,l_2)$.
\commentadd{
\begin{wrapfigure}{r}{1.4in}
	\centering
	\vspace{-0.3in}
    \includegraphics[width=1.4in]{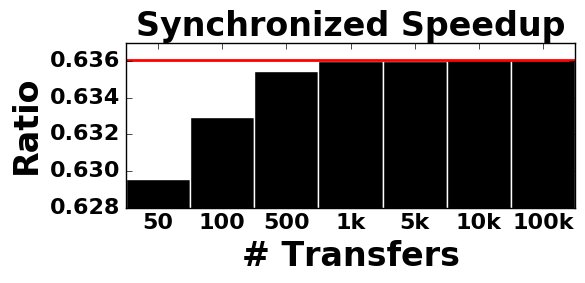}
	\caption{Empirical cost ratio versus the estimate}
  	\label{fig:bounds}
\end{wrapfigure}

Prior work~\cite{santalo2004integral} offered an estimate for the expected length of a transit path $C_{sg}$ in terms of the expected length of a line segment, $0.52$, in a randomized setting in an unit square. With the current estimate of $0.663$ for the maximum of two such randomly sampled line segments, it follows that, the expected makespan or maximum of distances cost will use this estimate.

Using this result, the synchronized cost ratio is stated in Equation~\ref{eq:synchronized-ratio} as
$$
\frac{C_{\rm dual}^{\rm sync}}{C_{\rm single}} \approx 
\frac{(c_{pd} + 0.66c_t)\frac{n}{2} + 4n\pi rc_t}{(c_{pd} + 0.52c_t)n}
$$
As a way to validate our asymptotic estimate, randomized trials were run with different number randomly sampled object transfer coordinates on an unit square.
% Fig \ref{fig:bounds} verifies empirically that the ratio of $\frac{C_{dual}^{sync}}{C_{single}}$ when $c_{pd}=0,r=0$, asymptotically converges to the calculated expected value of $0.636$. 
When $c_{pd}=0$ and $r=0$, the ratio of $\frac{C_{dual}^{sync}}{C_{single}}$ evaluates to $0.636$. Fig \ref{fig:bounds} verifies empirically that the ratio converges to the expected value as the number of transfers increases.
This indicates the asymptotic speedup of a synchronized dual arm solution for a makespan or maximum of distances cost metric.
}

\subsection{Smoothing}
\begin{figure}
	\centering
	\includegraphics[width=2.3in]{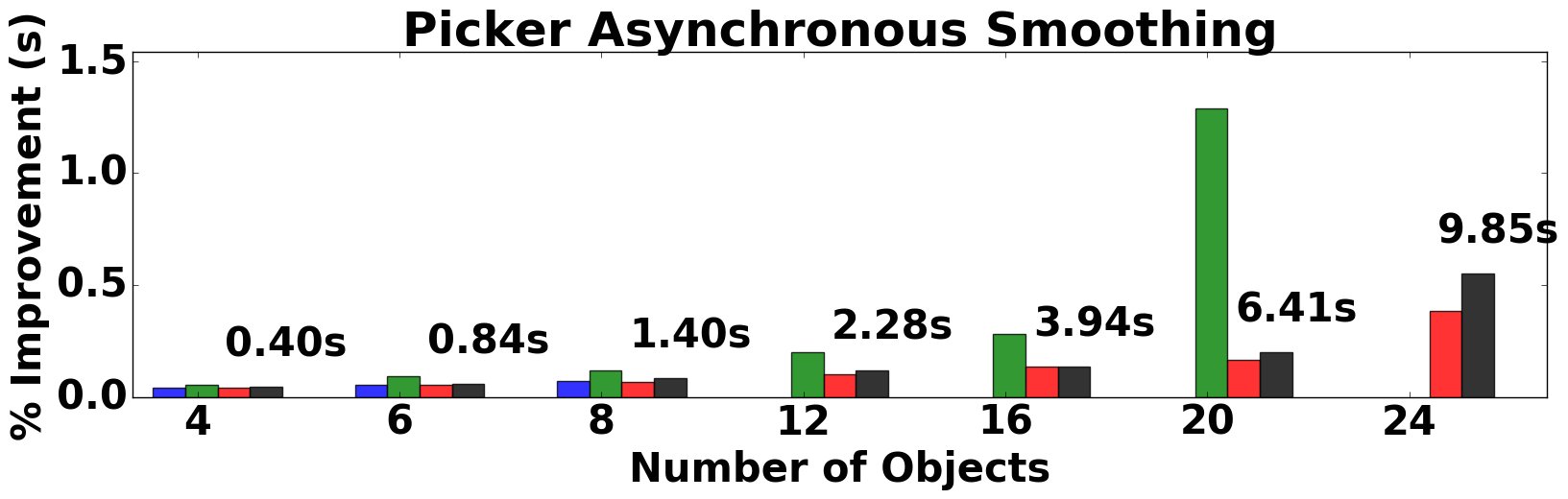}
	\includegraphics[width=2.3in]{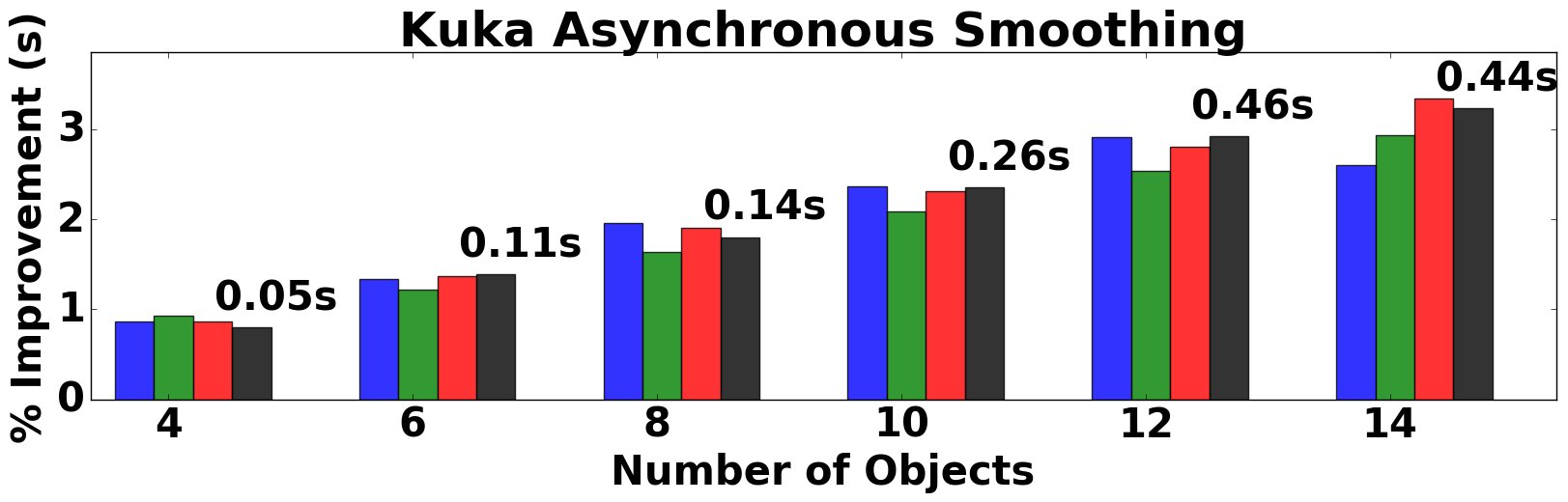}
	\caption{Smoothed solution improvement as a percentage of the original synchronized solution duration, and the time taken to smooth solutions obtained from \algo in seconds.}
	\label{fig:smoothing}
%     \vspace{-.3in}
\end{figure}

The result of the velocity tuning over the solution trajectories for the individual arms as a post-processing step is shown in Fig \ref{fig:smoothing}. The objective is to minimize any waits that might be a by-product of the synchronization. 
% The small improvements indicate that synchronization does not adversely affect the solution executions. 
The small \% improvements indicate that the asynchronous variants of the solutions discovered from the methods do not yield a big enough saving in execution time. 
Most of the improvement as a percentage of the original solution duration is not too high. On top of that, the time taken to smooth the solutions for \algo (overlaid on Fig \ref{fig:smoothing}) shows that it is often not beneficial.
In their largest problem instances the \kuka spent $ 0.44s $ of smoothing time to save $ 3.23s $ off the solution duration, while the picker spent $  9.84s $ to save $ 0.54s $.
This indicates that among the class of synchronized solutions discovered by the proposed algorithms, the analogous asynchronous variants do not seem to be drastically better. Moreover, smoothing does not improve the maximum of distances cost measure, but only reduces the solution duration. The theoretical bounds in the simple planar setups agree with the results in that the synchronization does not degrade the benefits of using 2 arms too much. It should be pointed out though that it needs to be studied further, whether these trends would hold for a class of algorithms that can solve the asynchronous dual-arm rearrangement problem in general setups. This is out of the scope of the current work.
 
% \kiril{What is the bottom line here? Is smoothing worthwhile? The y-axis in the plots is labeled as \% whereas the bars are labeled with seconds.}

% \vspace{-0.2in}

%####################################################################################
%####################################################################################
%####################################################################################
%####################################################################################
%####################################################################################
% {\small
% \bibliographystyle{spmpsci}
% \bibliography{bib/manip}}

% \end{document}

\end{document}